%% file: main.tex
\documentclass[10pt,twocolumn,letterpaper]{article}

\usepackage{iccv}
\usepackage{times}
\usepackage{epsfig}
\usepackage{graphicx}
\usepackage{amsmath}
\usepackage{amssymb}

\usepackage{graphicx}
\usepackage{booktabs}
\usepackage{enumitem}
\usepackage{soul}
\usepackage{xcolor}
\usepackage{caption}
\usepackage[pagebackref=true,breaklinks=true,letterpaper=true,colorlinks,bookmarks=false]{hyperref}
\usepackage[capitalize]{cleveref}
\crefname{section}{Sec.}{Secs.}
\Crefname{section}{Section}{Sections}
\Crefname{table}{Table}{Tables}
\crefname{table}{Tab.}{Tabs.}

\iccvfinalcopy %

\def\iccvPaperID{5361} %

\ificcvfinal\fi

\begin{document}

\newcommand{\mbf}[1]{\mathbf{#1}}
\newcommand{\AR}[1]{\textbf{\color{red} [AR:{#1}]}}
\newcommand{\JH}[1]{\textbf{\color{blue} [JH:{#1}]}}
\newcommand{\LZ}[1]{\textbf{\color{cyan} [LZ:{#1}]}}
\newcommand{\AK}[1]{\textbf{\color{brown} [AK:{#1}]}}
\newcommand{\NJ}[1]{\textbf{\color{magenta} [NJ:{#1}]}}
\newcommand{\AC}[1]{\textbf{\color{teal} [AC:{#1}]}}

\title{LANe : Lighting-Aware Neural Fields for Compositional Scene Synthesis}

\author{Akshay Krishnan*$^1$ \quad Amit Raj*$^1$ \quad Xianling Zhang$^2$   \quad 
Alexandra Carlson$^2$ \\  \quad Nathan Tseng$^2$ \quad Sandhya Sridhar$^2$  \quad Nikita Jaipuria$^2$ \quad James Hays$^1$\\
$^1$Georgia Institute of Technology \quad $^2$Ford Autonomy\\
{\tt\small $^1$\{akshay,amit.raj,hays\}@gatech.edu} \\
{\tt\small $^2$\{xzhan258, acarls66, ntseng3, ssridh38, njaipuri \}@ford.com}
}
\input{figures/fig-teaser}

\maketitle
\ificcvfinal\thispagestyle{empty}\fi

\newcommand\blfootnote[1]{%
  \begingroup
  \renewcommand\thefootnote{}\footnote{#1}%
  \addtocounter{footnote}{-1}%
  \endgroup
}

\begin{abstract}
   Neural fields\blfootnote{*equal contribution} have recently enjoyed great success in representing and rendering 3D  scenes. However, most state-of-the-art implicit representations model static or dynamic scenes as a whole, with minor variations. Existing work on learning disentangled world and object neural fields do not consider the problem of composing objects into different world neural fields in a lighting-aware manner. We present Lighting-Aware Neural Field (LANe) for the compositional synthesis of driving scenes in a physically consistent manner. Specifically, we learn a scene representation that disentangles the static background and transient elements into a world-NeRF and class-specific object-NeRFs to allow compositional synthesis of multiple objects in the scene. Furthermore, we explicitly designed both the world and object models to handle lighting variation, which allows us to compose objects into scenes with spatially varying lighting. This is achieved by constructing a light field of the scene and using it in conjunction with a learned shader to modulate the appearance of the object NeRFs. We demonstrate the performance of our model on a synthetic dataset of diverse lighting conditions rendered with the CARLA simulator, as well as a novel real-world dataset of cars collected at different times of the day. Our approach shows that it outperforms state-of-the-art compositional scene synthesis on the challenging dataset setup, via composing object-NeRFs learned from one scene into an entirely different scene whilst still respecting the lighting variations in the novel scene. For more results, please visit our project website \url{https://lane-composition.github.io/}.
\end{abstract}
\input{sections/intro}

\input{sections/related}

\input{sections/approach}
\input{sections/experiments}

\input{sections/discussions}
\input{sections/conclusion}

{\small
\bibliographystyle{ieee_fullname}
\bibliography{egbib}
}

\input{supplm}

\end{document}

%% file: figures/fig-teaser.tex
\twocolumn[{%
\renewcommand\twocolumn[1][]{#1}%
\vspace{-3em}
\maketitle
\thispagestyle{empty}
\vspace{-3em}
\begin{center}
    \centering
    \includegraphics[width=\linewidth]{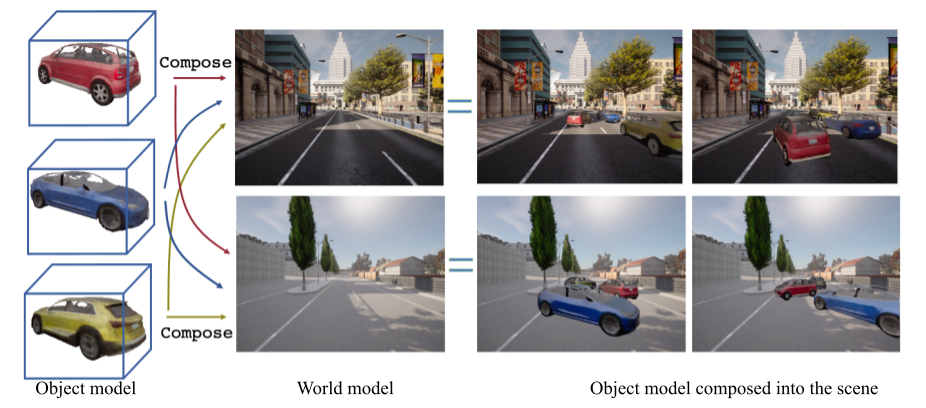}%
    \captionof{figure}
    {We present Lighting-Aware Neural Fields (LANe) for compositional scene synthesis. With the disentanglement of a class-specific object model (Column 1) and the learned world model (Column 2), LANe can arbitrarily compose objects into different scenes (Column 3 and 4). Our novel light field modulated object model can be composed into  scenes in a lighting-aware manner. The figure above shows the same world model used as background scenes on each row, and object models composed into them in arbitrary poses under different lighting conditions. Note that the composed objects are shaded appropriately based on the local lighting condition at the placed location, which shows LANe's spatially varying lighting-aware compositional synthesis capabilities. 
    }
    \label{fig:teaser}
\end{center}%
}]

%% file: sections/intro.tex
\newenvironment{packed_enum}{
\begin{enumerate}
  \setlength{\itemsep}{1pt}
  \setlength{\parskip}{0pt}
  \setlength{\parsep}{0pt}
}{\end{enumerate}}

\vspace{-0.2in}
\section{Introduction}

Controllable synthesis of a wide variety of road scenes is of particular interest for training and validating autonomous driving perception systems. Specifically, tasks such as re-simulation and synthesis of rare scenarios require control of a wide variety of 3D scene properties. Precise and controllable 3D scene generation has been a long-standing challenge in computer vision. While there has been significant progress in developing traditional photo-realistic rendering engines \cite{CARLA2017}, they still suffer from the synthetic to real domain gap. Furthermore, significant effort is expended to author photo-realistic 3D assets. This often requires considerable artistic skill and expertise, not to mention cost. In contrast, existing model-based simulators \cite{cabon2020virtual, gaidon2016virtual} provide controllable synthesis of scenes and images without the need for digital artists. However, they still suffer from a distribution shift with respect to real world images. 

Neural Radiance Fields (NeRFs), a state-of-the-art neural rendering technique, offer a promising solution to this problem. NeRFs have been leveraged for learning 3D scene representations for both simple synthetic scenes as well as complex, in-the-wild, multi-view image datasets \cite{mildenhall2020nerf, yu2021artificial, raj2020pva, verbin2022ref, zhang2020nerf++, gao2021dynamic}. However, most NeRF research addresses the problem of modelling a 3D scene as a whole, which does not allow for composition.

As shown in Fig.~\ref{fig:teaser}, we investigate the task of inserting cars into driving scenes, where both the scene and the car are represented by 3D neural representations learned from 2D image sequences. Our scene and object representations are lighting aware. This allows us to insert objects in novel poses and novel scenes, while modulating their appearance to be locally consistent with the lighting of the scene.

Our work builds upon recent works that have also addressed the problem of compositional scene modeling with NeRFs by separating scene and object models. Neural Scene Graphs\cite{ost2021neural} learned a scene graph to model the driving scene, with world and object models represented by implicit representations, but were limited to the same source lighting conditions.
Panoptic neural fields\cite{kundu2020appearance} includes representations like semantic and instance segmentation allowing for more fine grained addition and removal of objects in the scene. Several other works such as \cite{wu2022object,wu2022d}, representing dynamic scenes present frameworks to separate the static and dynamic components. Most of these frameworks ground the dynamic component either to a time variable or to a learnt latent variable. This assumption restricts the variation in object compositions that can be performed with respect to scene. However, in all such previous works, the composition is not lighting-aware (i.e the object's appearance is inconsistent with the scene's lighting conditions). 

The limitations of prior work \cite{ost2021neural, kundu2020appearance, wu2022object, wu2022d} fail to produce realistic results by lacking lighting-aware composability. We directly address these shortcomings with our novel approach for scene modeling and object insertion in a lighting-aware manner, without explicitly modelling the materials of the objects and the scenes. To summarize, the contributions of our work are as follows,

\begin{packed_enum}
    \item We present a 3D neural scene representation that represents the scene with a world-NeRF for the background and class-specific object-NeRFs for the dynamic elements, both of which are lighting dependent. 
    \item We propose a novel approach to modulate the color of the rendered objects in unseen poses and scenes, by augmenting the scene with a spatially varying light field and the object with a lighting-dependent shader. 
\end{packed_enum}

%% file: sections/related.tex
\section{Related Works}

\subsection{2D methods}
2D approaches address the problem of compositional synthesis by alpha compositing different elements of the scene in a layered manner. Omnimatte\cite{lu2021omnimatte} separates static and dynamic elements of a video and associates correlated dynamic effects to the corresponding element. Layered neural rendering separates out the scene and dynamic elements. Alhaija et al.\cite{alhaija2018geometric} uses digital assets rendered onto a scene followed by a network to harmonize the inserted object using an adversarial loss. Whilst all these approaches demonstrate the state-of-the-art performance on scene synthesis and composition, they lack 3D scene understanding and the level of controllability of scene manipulation. With 3D geometry and lighting awareness, our approach models both the dynamic and static elements of the scene as implicit representations, allowing for controllable composition of elements into arbitrary locations.

\subsection{Explicit 3D methods}
Several approaches \cite{raj2020anr, ZhangTSBJ22, Granskog2020} have since used explicit 3D models for compositional scene synthesis. Specifically, Raj et al.\cite{raj2020anr} use a mesh proxy to represent the dynamic human avatar and compose it into arbitrary scene using deferred neural rendering. Neural light field for object composition demonstrate it with explicit mesh while estimating lighting of real scenes. From a single 2D image, SIMBAR \cite{ZhangTSBJ22} models the scene as a 3D mesh to explicitly represent scene geometry, followed with shadow refinement network to produce realistic shadows. Granskog et al \cite{Granskog2020} propose a technique to compose neural scene representation for shading inference, which explicitly distentangles lighting, material, and geometric information using illumination buffers. These discussed explicit 3D methods lack object-aware composition capability with the scene.

\subsection{Implicit 3D methods}
Several recent methods study the problem of composing scene elements with implicit representation\cite{mildenhall2020nerf,sitzmann2020implicit,yu2021artificial, yu2021plenoctrees, yu2021plenoxels,raj2022dracon}, which have gained a great success in modeling scenes. Particularly, ObjectNeRF \cite{wu2022object} consists of an object model that is used to represents parts of the scene other than the background, and a scene model that is responsible for recomposing the decomposed objects to the scene. Neural Scene Graphs (NSG) \cite{ost2021neural} models driving scenes with a world model and object models of learned scene graph representation, which encodes object transformation and radiance. Panoptic Neural Fields \cite{kundu2022panoptic} extends NSG to predicts panoptic-radiance field that encodes color, density and semantic segmentation labels of the objects in the scene. However, most of the models do not model effect of lighting changes on the scene or objects. 

\subsection{Lighting-aware representations}
Several approaches \cite{boss2021nerd,srinivasan2021nerv,zhang2021nerfactor,boss2021neural} have worked on lighting-aware manipulation for single objects or scenes. Particularly, NeRF-OSR\cite{rudnev2022nerf} leveraged scene geometry surface property modeling to account for outdoor scenes captured under varying illumination, but it was restricted mostly to static building architecture. NeRFactor\cite{zhang2021nerfactor} modeled the lighting effect on objects using a BRDF represented by an implicit representation. Zhang et al. \cite{zhang2021ners} proposed to learn object surfaces and use the Phong shading model \cite{Phong75Shading} to capture lighting variations. However, none of these methods address the interactions of objects with their surrounding world, namely, for NeRD, it assumes light sources are at infinity and needs observations of the object at a certain location to build an environment map, or an environment map of the scene at the particular pose that needs to be rendered. Such an approach has the limitation that it is computationally intensive to compute environment maps at each pose.

For compositional synthesis, prior models have not considered whether incident lighting is dependent on the target location for composition within the overall world model. This is an important question is not addressed by work before LANe. Our work is similar in spirit to OSF\cite{OSF2020}, which is also evaluated primarily on synthetic scenes; however, we note that OSF only primarily works on point light sources in indoor scenes under known lighting conditions. In contrast, our method uses lighting information gathered directly from the scene, and tackles the much harder problem of outdoor scenes.
Furthermore, in contrast to methods that model the material properties explicitly, our approach learns the lighting effect as a multiplicative term on top of the learnt radiance without the need to model accurate BRDF material properties. Furthermore, our approach eschews the expensive requirements to compute a lighting representation at each rendered pose, and can interpolate between training poses. 

Generating data using continuous composition with spatially varying lighting in outdoor driving scenes, it allows us to facilitate the data need of autonomous driving perception systems. This is the main differentiator between our approach and the existing methods. LANe is able to compose dynamic moving objects(vehicles) and continuously changing outdoor environments in a lighting-aware manner.

%% file: sections/approach.tex
\section{Approach}
\input{figures/fig-architecture.tex}

\input{figures/fig-comparison.tex}

\subsection{Preliminaries} 
We base our representations on NeRFs\cite{mildenhall2020nerf, barron2021mip}, which use MLPs to learn a 3D volumetric model from posed images. Specifically, given a set of images $\{I_i\}_{i=1}^k$ with known camera locations $\{\mathbf{o_i}\}_{i=1}^k$, we learn a scene representation $\mathcal{N}:\mathbb{R}^n \rightarrow \mathbb{R}^4$ such that pixel value observed along a particular direction $\mathbf{d}$ is obtained by casting a ray $r(t) = \mathbf{o} + t \mathbf{d}$ and performing volumetric integration along the ray as follows: 
\begin{equation}
    \mathbf{C} = \int_{t_{near}}^{t_{far}} T(t) \sigma(r(t)) c(r(t)) dt
\end{equation}

where $(c, \sigma)$ are the outgoing radiance and density at a 3D point modelled by the NeRF $(\mathcal{N})$, $(t_{near}, t_{far})$
 are the near and far plane boundaries along the ray, and $T$ is the accumulated transmittance along the ray, given by $T(t) = \exp( -\int_{t_{near}}^t (\sigma(r(s))ds)  $

In practice, the discrete version of the integration is performed by quadrature approximation during volume rendering as given by \cite{max1995optical}.

\subsection{Overview}
Following \cite{ost2021neural,kundu2022panoptic}, as illustrated in Fig.~\ref{fig:framework}, we decompose the scene into a world-NeRF $\mathcal{N}_{world}$ (Sec.~\ref{sec:wm}) and an object-NeRF $\mathcal{N}_{obj}$ (Sec.~\ref{sec:om}) to represent the static and dynamic components respectively. Additionally, we train both our object and world models in a lighting aware manner, under multiple lighting conditions, to disentangle geometry and albedo from lighting effects. Particularly, our datasets comprise a set of images of multiple scenes $\{I_i^{(l_j)}\}_{i=1}^K$ under lighting conditions $l_j \in \mathcal{L}$. 

For world-NeRFs in driving scenes, we assume a single light source at infinity and model the lighting effects on the world as a function of azimuth and elevation angle $(\theta, \phi)$. As indicated in Fig.~\ref{fig:varying_lighting}, the scene model produces spatially-varying intermediate lighting features $f_d$ that are fed to the object-NeRF to condition the object's appearance on spatially-varying lighting cues. 

\subsection{World-NeRF}
\label{sec:wm}

The scene is modeled by a lighting agnostic network $\mathcal{N}_{world}$ and lighting aware Neural Light field $\mathcal{N}_{light}$ \cite{sitzmann2021light}.
Particularly,
\begin{equation}
\mathbf{c},\sigma= \mathcal{N}_{world}(\mathbf{x})
\end{equation}
\begin{equation}
\label{eq:world-light}
\mathbf{c_{lf}} = \mathcal{N}_{light}(\mathbf{o},\mathbf{d}; \mathbf{f})
\end{equation}
where $\mathbf{x}$ is a sample along the ray, $\mathbf{o}$ and $\mathbf{d}$ are ray origins and directions, and $\mathbf{f}$ is a latent feature that parameterizes the scene lighting. This could either be learned or set from physical lighting parameters (sun azimuth and elevation). 

Since $N_{world}$ only takes the spatial position as input but is trained across different lighting conditions, it essentially learns an average color across all lighting variations. The final color of the world scene given by a multiplicative effect of the lighting agnostic scene geometry and the lighting aware scene model.  Particularly, we use $\mathbf{f}$, the light-field latent, to learn a lighting-specific multiplier.
\begin{equation}
    \mathbf{\tilde{c}} = \mathbf{c} * \mathcal{N}_{ws}(\mathbf{f})
\end{equation}
Where $\mathcal{N}_{ws}$ represents a shader function that learns a multiplicative factor to obtain the lighting specific scene radiance from the lighting agnostic color. The light field outputs $\mathbf{c_{lf}}$ are also used to shade the inserted object models are described in \ref{sec:om}.

\subsection{Object-NeRF}
\label{sec:om}
The object model, similar to the world model, has two components: a scene-agnostic representation for density and color (albedo) $\mathcal{N}_{obj}$, and a scene-dependent shader for radiance $\mathcal{N}_{shading}$. The coordinate inputs to the models $\mathcal{N}_{obj}$ and $\mathcal{N}_{shading}$ are represented in normalized object coordinate frames\cite{wang2019normalized}. Their weights are shared across object instances of the same semantic class (cars) with different colors and shapes, by using instance-specific shape and color codes inspired by \cite{jang2021codenerf,liu2021editing}. 

Specifically, the scene-agnostic representation is modeled as: 

\begin{equation}
 \mathbf{c}, \sigma = \mathcal{N}_{obj}(\phi(\mathbf{x}_{object}))
\end{equation}

$\mathbf{c}$ above is the lighting-agnostic radiance of the object. The lighting conditioned radiance is obtained by multiplying $c$ with a shading coefficient $s_{car}$ predicted by $\mathcal{N}_{shading}$. 

\vspace{-0.2cm}
\begin{equation}
\mathcal{\mathbf{\tilde{c}}} = \mathbf{c} * s_{car} 
\end{equation}

We then use $\bar{\mathbf{c}}$ during volumetric rendering as in \cite{mildenhall2020nerf}. $\phi$ is the position encoding applied to its inputs, as is standard practice for neural fields. We explore different shader architectures optimized for two different downstream applications: for composing objects into new locations within the same scene, and for composing into new scenes. These reflect two methods for feeding information from the world representation to the object shading representation. 

\vspace{-0.2cm}
\subsubsection{Composing into known scenes}
\label{sec:known}

For composing an object model into new locations in scenes where it has already been observed, the input coordinates has been fed in the global frame in addition to the coordinates in object coordinate information to represent a lighting aware object model. Specifically, given the 3D bounding box of the car with parameters $\mathbf{R_{car}}$ and $t_{car}$, we transform the rays cast into the scene into object coordinate system as follows. 

\begin{equation}
    \mathbf{x}_{object} = [\mathbf{R}_{car}|t_{car}] \mathbf{x}_{scene}
\end{equation}

Then the shading network is modelled as:

\begin{equation}
s_{car} = \mathcal{N}_{shading}(\phi(\mathbf{x}_{object}), \phi(\mathbf{x}_{scene}), \mathbf{\phi(R_{car})},  \mathbf{\tau})
\end{equation}

where $\mathbf{\tau}$ is a learnable scene specification that allows us to share weights for $\mathcal{N}_{shading}$ between scenes. The shading network learns to shade the point  at $\mathbf{x}_{object}$ differently based on its global state ($\mathbf{x}_{scene}, \mathbf{R_{car}}$). 

\subsubsection{Composing into unknown scenes}
\label{sec:unknown}

To insert our object model into scenes where the object has not been observed during training, we need a shading model that does not specifically memorize the scene. To this end, we learn a generalizable shader that uses the light-field of the target scene $\mathcal{N}_{light}$ to compute $s_{car}$
Recollect that the rendering equation for the output radiance at a point can be written as follows:

\begin{equation}
L_{out}(\omega_o) = \int_{\omega_i \in \Omega} f(\omega_i,\omega_o) L_{in}(\omega_i) (\mathbf{\hat{n}}.\mathbf{\hat{d}}) d\omega
\end{equation}

Where, $\omega_i$ and $\omega_o$ are incoming and outgoing ray directions respectively, $\mathbf{\hat{n}}$ is the normal calculated at the surface and $\mathbf{\hat{d}}=-\omega_o$ is the viewing direction.
Here, we model $L_{in}(\omega_i)$ with the scene light-field $\mathcal{N}_{light}$. We also assume a Lambertian model of the object, which reduces $f(\omega_o,\omega_i)$ to a constant. We approximate the integral by a weighted sum. In particular, for each point on the surface of the object to be rendered $\mathbf{p}$, we cast secondary rays to evaluate the incoming light as 
\begin{equation}
    \mathbf{l_d} = \mathcal{N}_{light}(\mathbf{p},\mathbf{d}) 
\end{equation}

Since the normals from the density fields can be noisy, we use attention layers, with the local car coordinates as queries, and the incident lighting values $\mathbf{l_d}$ along with directions $\mathbf{d}$ as keys and values, to summarize the incident lighting at each point in a feature $\mathbf{\tilde{f_l}}$. More details about the attention mechanism are provided in supplementary.

This accumulated feature $\mathbf{\tilde{f_l}}$ is fed into a shading MLP along with the coordinates of the point in the local car coordinates $\mathbf{x_{object}}$ to predict a shading value. 
\begin{equation}
    s_{car} = \mathcal{N}_{shading}(\mathbf{\tilde{f_l}},\mathbf{p})
\end{equation}

\subsection{Training}
We train the object-NeRF and world-NeRF with the following objectives:

\textbf{Photo-metric loss}: This encourages the rendered pixel to match the color of the ground truth pixel color $\hat{\mathbf{C}}$. 
\begin{equation}
    \mathcal{L}_{p}= ||\mathbf{C}(\mathbf{r})-\hat{\mathbf{C}}||
\end{equation}

\textbf{Mask-loss} : We find that the mask loss is necessary to separate out the objects from the scene. 
\begin{equation}
    \mathcal{L}_{mask}=||M(\mathbf{r})-\hat{M}||
\end{equation}
where $M(\mathbf{r})$ and $\hat{M}$ represents the accumulated alpha value along a ray $r$ and ground truth mask respectively.

\textbf{Depth guidance} : In datasets where we have access to depth/lidar information, we leverage depth for the training rays in the world-NeRF as in  \cite{deng2022depth,rematas2022urban}.
\begin{equation}
    \mathcal{L}_{depth}=||z(\mathbf{r})-\hat{z}||
\end{equation}
\input{figures/fig-ablation}

Furthermore, we find that, coarse to fine grained optimization helps in improving the quality of the learnt object model. 

The loss for the world model is then given as follows:
\begin{equation}
\mathcal{L}_{world} = \lambda_{p} \mathcal{L}_p^{(world)} + \lambda_{depth} \mathcal{L}_{depth}^{(world)}
\end{equation}

And the corresponding object model is given by:
\begin{equation}
\mathcal{L}_{object} = \lambda_{p} \mathcal{L}_p^{(object)} + \lambda_{mask} \mathcal{L}_{mask}^{(object)}
\end{equation}

The lighting-agnostic object and world models can be trained independently as we have annotations for the static and dynamic components of the scene. The training of the object model uses the light field component of the world model.

\input{figures/fig-ablaiton_model.tex}

%% file: figures/fig-architecture.tex
\begin{figure*}[h!]
    \centering
    \includegraphics[width=\textwidth]{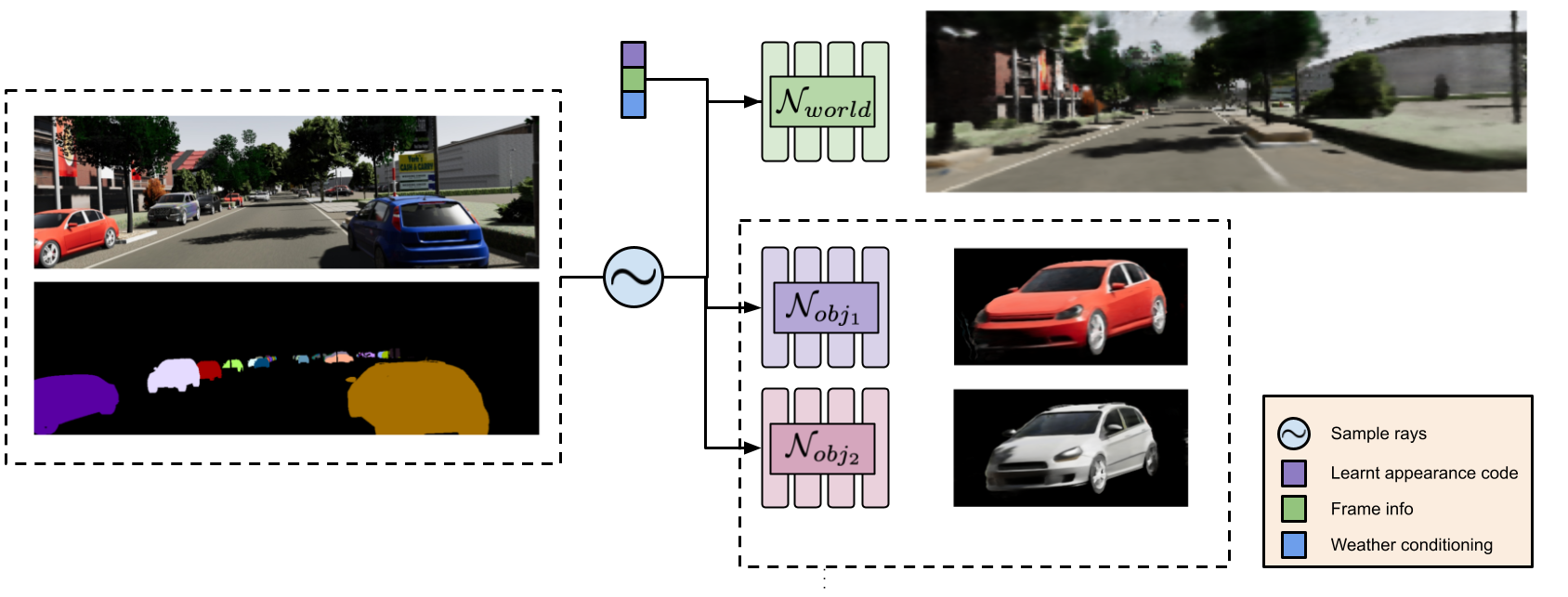}\\
    \includegraphics[width=\textwidth]{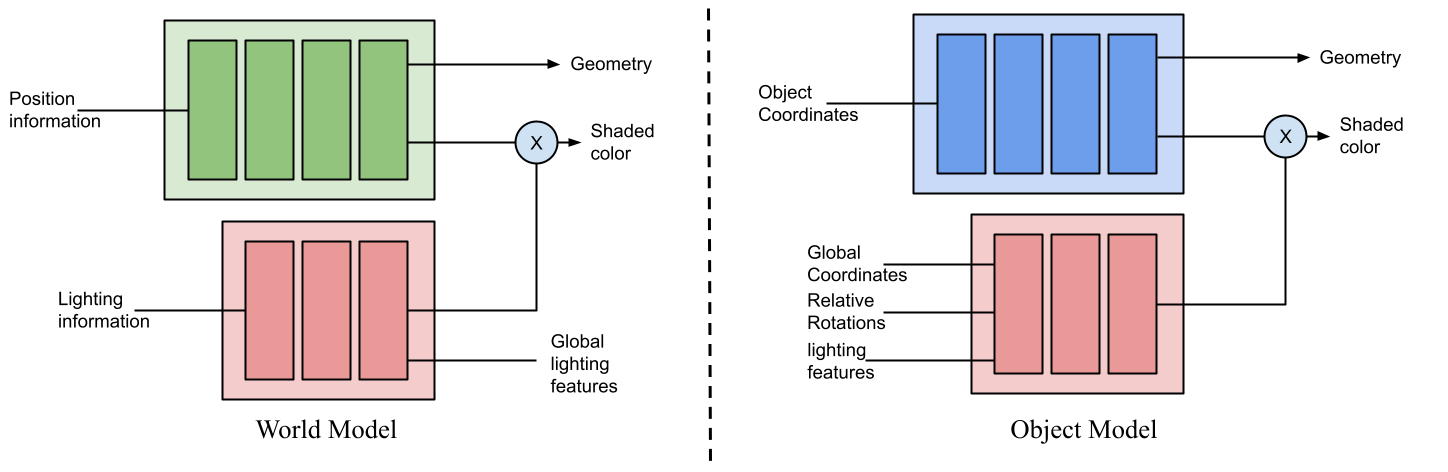}
    \caption{Overview of the proposed approach. We model the scene with a seperate world-NeRF, which lighting-aware by training on the same scene under different lighting conditions. 
    (Sec.~\ref{sec:wm}) and a class specific object-NeRF, which  use information from the scene-NeRF to train the object NeRF. $\mathcal{N}_{obj}$(Sec.~\ref{sec:om}) 
    } 
    \label{fig:framework}
\end{figure*}

%% file: figures/fig-comparison.tex
\begin{figure*}[h!]
    \centering
    \includegraphics[width=\textwidth]{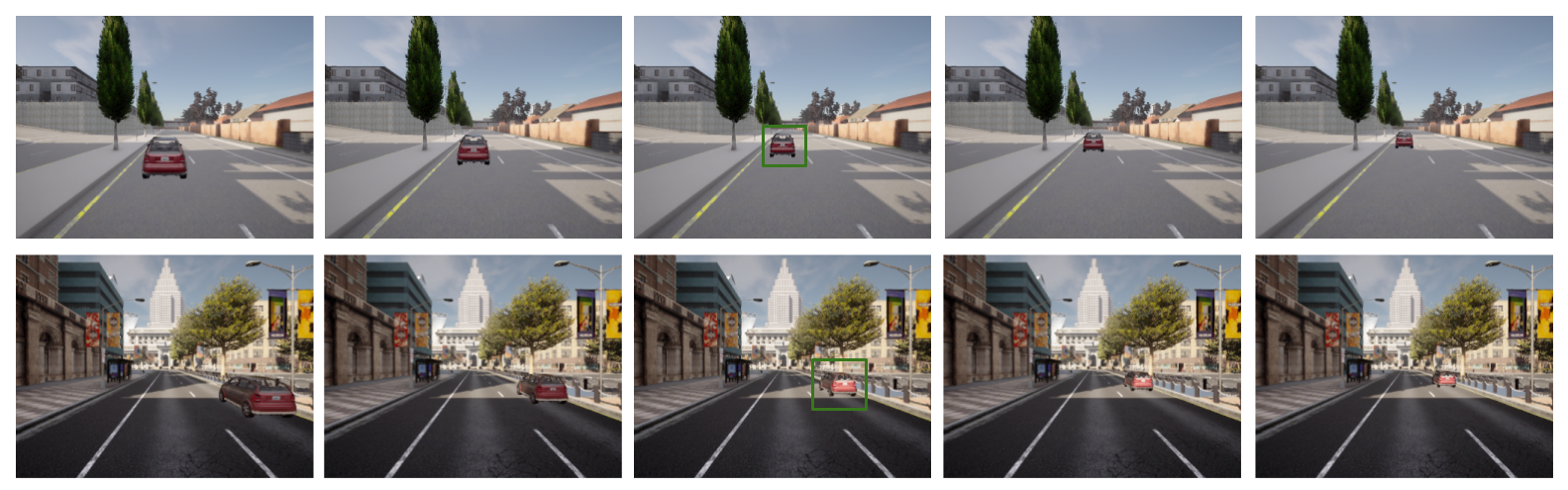}
    \caption{LANe can synthesize scenes with object models that respect spatially varying lighting. This figure shows the object model moving through the scene with spatially varying lighting, we observe that the object gets brighter as it enters a region of light from a region of shadow. 
    } 
    \label{fig:varying_lighting}
\end{figure*}

%% file: figures/fig-ablation.tex
\begin{figure}[h!]
    \centering
    \includegraphics[width=\linewidth]{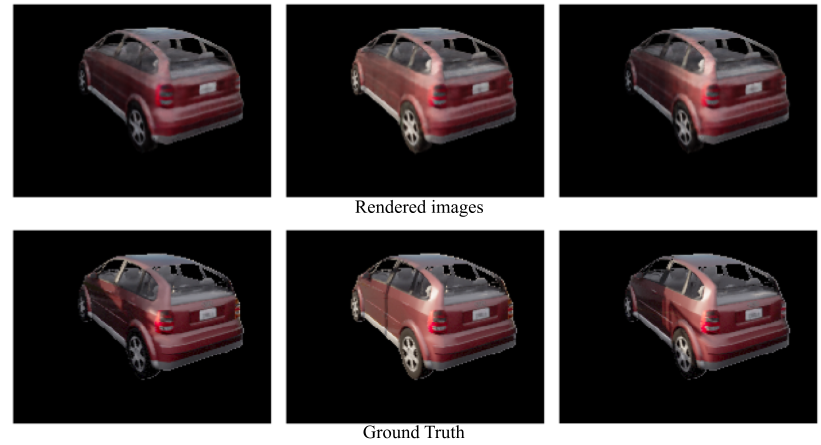}
    \caption{Our Object model rendered under different lighting conditions {\color{blue}(Top row)} and corresponding ground {\color{blue}(Bottom row)}. We observe that our multiplicative model captures spatially varying lighting effects despite not explicitly modeling normals.}
    \label{fig:lighting}
\end{figure}

%% file: figures/fig-ablaiton_model.tex
\begin{figure*}[h!]
    \centering
    \includegraphics[height=8cm]{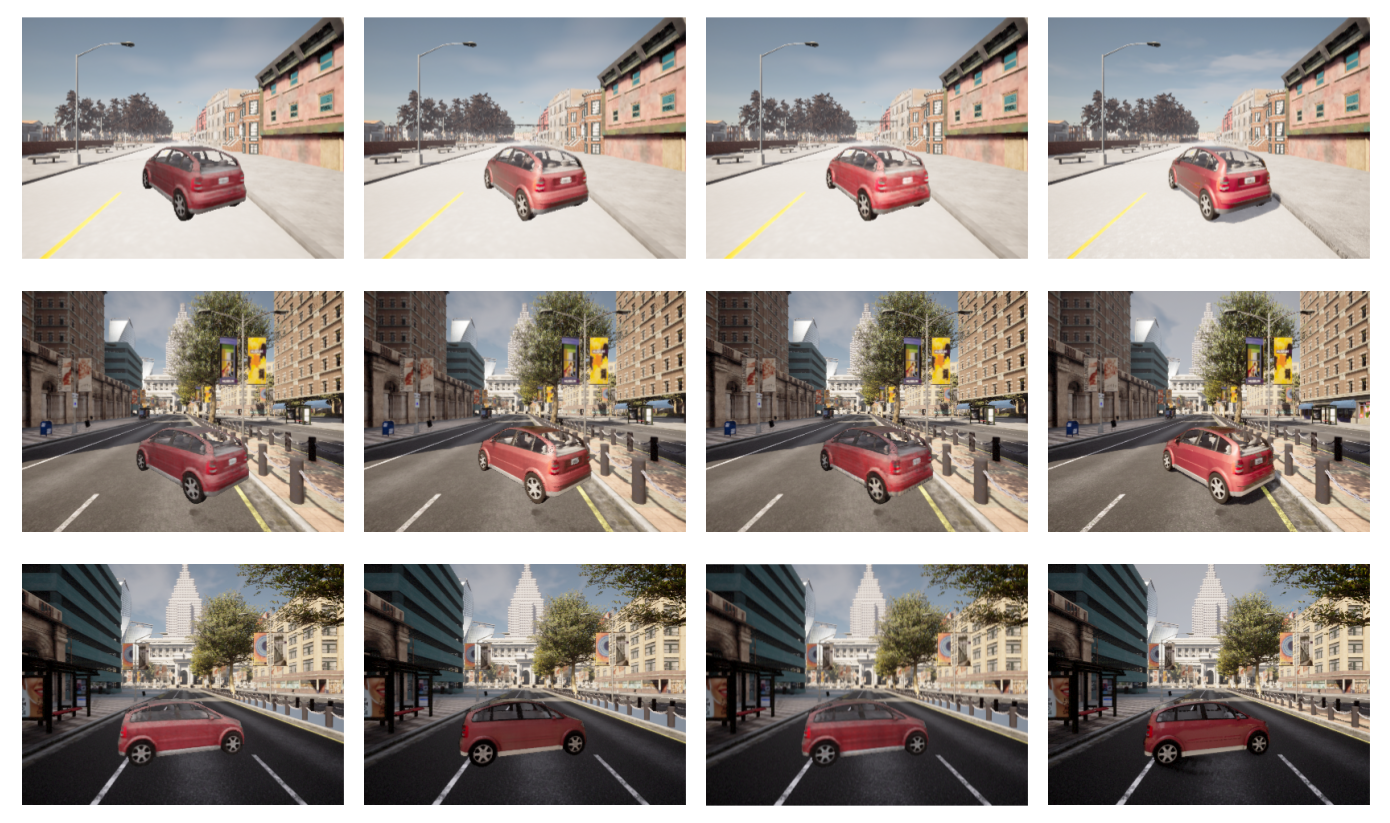}    
    \caption{A comparison of our models for lighting aware object-composition. Column 1: new object model inserted but unshaded; Column 2: local-global network; Column 3: light field conditioned model; Column 4: ground truth.}
    \label{fig:object_composition}
\end{figure*}

%% file: sections/experiments.tex
\section{Experiments}

Our experiments evaluate the quality of images rendered when composing LANe models into unseen poses (Fig.~\ref{fig:varying_lighting}) within both seen and unseen worlds. For seen environments, we  report metrics for both the local-global coordinate shader (Sec.~\ref{sec:known}) as well as the light-field conditioned shader (Sec.~\ref{sec:unknown}), although the main benefit of the latter is its ability to generalize to unseen environments. 

\subsection{Architecture details}
\label{sec:arch}

\textbf{World NeRFs} Our world NeRF $\mathcal{N}_{world}$ is represented as a standard NeRF with 8 MLP layers and a parallel branch with 4 MLP layers $\mathcal{N}_{ws}$ to control the lighting of the associated scene. We use a similar 8-layer MLP to learn a light field network $\mathcal{N}_{light}$ for the scene.

\textbf{Object NeRFs} Our base object NeRF $\mathcal{N}_{object}$ follows a similar 8-layer MLP architecture. 

\textbf{Object Shader} In known scenes, the shader network is simply another MLP that accepts the global coordinates and the orientation of the car as a quaternion. The  shader network for unknown scenes is modeled as an attention-conditioned MLP with the local coordinate attending over sampled light field directions and values.

\subsection{Datasets}

\textbf{Synthetic CARLA dataset}
We use the CARLA simulator \cite{dosovitskiy2017carla} to render images of multiple urban scenes under varying lighting conditions with vehicles in different poses in the scene. We render 8 scenes under different lighting conditions with 5 different car instances observed in 40 different locations in each scene. We divide the scenes into 6 training scenes and 2 test scenes for evaluating composition into novel scenes. Within each scene, we also hold-out 20\% of the locations, to evaluate composition in known scenes. Unlike many other NeRF datasets, our camera orientation does not vary significantly, to be representative of real-world car data. 

\textbf{Real world dataset}
To evaluate the applicability of our approach to real world images, we collect a real world multi-view dataset of 4 different cars at 10 different times of the day. Each instance comprises videos from a handheld mobile camera revolved around the car. ~100 frames are extracted from each video to train object and world models. We estimate the camera poses using COLMAP \cite{schoenberger2016sfm, schonberger2016structure}, predict 2D instance segmentation masks using an off-the-shelf model \cite{maskrcnn2017he}, and manually label approximate 3D bounding boxes from COLMAP reconstruction.

\input{tables/tab-total}

\subsection{Baselines}

\textbf{NeRF} We compare our method against lighting agnostic compositional 3D scene synthesis methods \cite{ost2021neural, kundu2022panoptic} by using a vanilla NeRF model as the object representation. Although they are not directly applicable to composable representations of dynamic objects, we explore the use of relightable NeRF methods \cite{boss2021nerd} in the supplementary material. 

\subsection{Known worlds lighting-aware composition}

In this experiment (shown in Fig.~\ref{fig:lighting}), we evaluate the quality of composing LANe object models into unseen poses in 3 environments with both globally as well as locally varying lighting conditions. These world environments have been used during training. Our results are reported in Tab.~\ref{tab:comparison}. We find that when composing into environments seen during training, the local-global architecture greatly outperforms other approaches. 

\subsection{Unseen worlds lighting-aware composition}

Only the light-field conditioned shading architecture (LANe-LF) is suitable for composition into unseen world models. In Fig.~\ref{fig:object_composition}, we compare this against the vanilla NeRF model and the local-global architecture (both of which have used this environment during training) in Table \ref{tab:ablation_sg}. We find that while LANe-LF is clearly better than a lighting-agnostic NeRF model, it is still slightly worse, but comparable in quality to a local-global model which was trained on this world.

\subsection{Real-world lighting aware composition}
\label{sec:real-composition}
\input{figures/fig-real}
\input{tables/tab-real}

For real scenes, the lighting aware object models are trained along with the world light fields, and composed into seen and unseen scenes. The results are as shown in Fig. \ref{fig:real}. Note that the unshaded lighting-agnostic object models still have some lighting artifacts in regions that were well-lit. The shader compensates for this, and brightens specular and well-lit regions on the car, while darkening regions that are not well-lit. The quantitative evaluation on the composition of the lighting-aware object model on held-out views and report findings are listed in Tab. \ref{tab:real-known}. The metrics show that the model shades the object in a lighting-consistent manner even in such scenes with challenging lighting and reflections. 

We find that training the attention-based shader architecture from Sec. \ref{sec:unknown} is challenging when objects are observed in only 7 lighting conditions. We therefore use global lighting features with the object shader, which captures inter-scene lighting information, but limits the spatial variance of object's appearance within the scene. 

\subsection{Ablations}

\textbf{Local-Global NeRFs for a single vs multiple objects} 
From Tab.~\ref{tab:comparison}, we observe training instance-specific lighting-aware models performs better than a shared model conditioned on an instance-specific latent code. Using a latent space to model multiple instances can however still be useful in practical settings where a single instance may not be observed in several parts of the scene. The multi-instance model would be able to share lighting information across instances.
\vspace{-0.1in}
\subsubsection{Local-Global Net architecture} 
\vspace{-0.1in}
Our proposed local-global net architecture uses the local coordinates for the density branch and the global coordinates for the radiance branch. We find that this separation of global and local coordinates is crucial, and that merely using the global coordinates as an input to the density branch along with the local coordinates results in the network learning incorrect densities, especially without a mask loss. 

We train a model with a global conditioning code instead of spatially varying local code as explained in our model.

\input{tables/tab-comparison.tex}

%% file: tables/tab-total.tex
\newcommand{\tn}[2]{ $#1$ {\color{blue} \tiny $\pm #2$} }
\newcommand{\tnbf}[2]{ $\mathbf{#1}$ {\color{blue} \tiny $\pm \mathbf{#2}$} }

\begin{table*}[h]
\small
    \centering
    \begin{tabular}{l|ccc | ccc | ccc}
    &\multicolumn{3}{c}{Scene 1} & \multicolumn{3}{c}{Scene 2} & \multicolumn{3}{c}{Scene 3} \\
    \specialrule{0.12em}{0.05em}{0.05em}
     & SSIM $\uparrow$ & PSNR $\uparrow$ & LPIPS $\downarrow$ & SSIM $\uparrow$ & PSNR $\uparrow$ & LPIPS $\downarrow$ & SSIM $\uparrow$ & PSNR $\uparrow$ & LPIPS $\downarrow$ \\
     \specialrule{0.12em}{0.05em}{0.05em}
    NeRF &  $0.783$ & \tn{17.696}{0.94} & $0.176$ & $0.798$ & \tn{22.115}{0.70}  & $0.188$ & $0.756$ & \tn{19.064}{2.14} & $0.196$ \\
    LANe (Single) &  $\mathbf{0.965}$ & \tnbf{27.754}{0.99}  & $0.151$ & $\mathbf{0.947}$ & \tnbf{26.304}{2.91}  & $\mathbf{0.062}$  & $\mathbf{0.939}$ & \tnbf{25.89}{5.19} & $\mathbf{0.059}$  \\
    LANe (Multiple) &   $0.837$ & \tn{22.548}{2.77}  & $\mathbf{0.077}$ & $0.857$ & \tn{25.353}{1.95} & $0.141$ & $0.739$ & \tn{19.791}{6.077}  & $0.213$\\
    LANe (LF) &  $0.864$ & \tn{22.161}{2.10} & $0.096$ & $0.868$ & \tn{24.353}{2.02} & $0.103$ & $0.862$ & \tn{22.017}{2.21} & $0.097$  
    \end{tabular}
    \caption{A comparison of image quality when composing object models into known world models.}
    \label{tab:comparison}
\end{table*}

%% file: figures/fig-real.tex
\begin{figure}[h!]
    \centering
    \includegraphics[width=\linewidth]{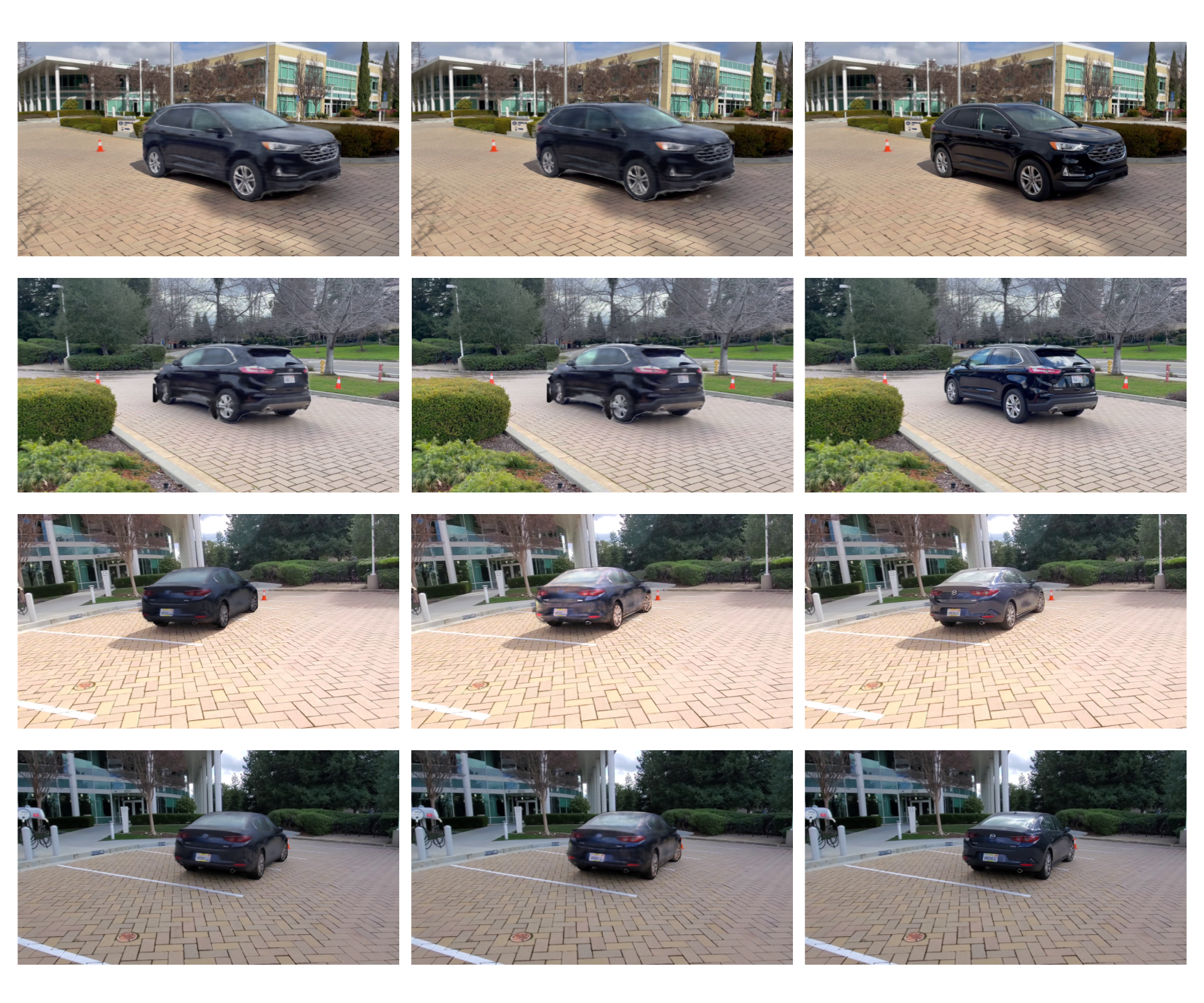}
    \caption{The lighting aware object model trained on real data composed into scenes with different lighting. Column 1: Lighting agnostic object model, Column 2: Object model with lighting-aware shading, Column 3: Ground truth.}
    \label{fig:real}
\end{figure}

%% file: tables/tab-real.tex
\begin{table}[h!]
\vspace{-0.2em}
\footnotesize
    \centering
    \begin{tabular}{l|ccc}
    \hline
    Method & SSIM $\uparrow$ & PSNR $\uparrow$ & LPIPS $\downarrow$ \\
    \hline
    NeRF only & $0.928$ & 24.390 & $0.081$  \\
    LANe (LF) & $0.937$ & 26.637 & $ 0.057 $  \\
    \hline
    \end{tabular}
    \vspace{-1mm}
    \caption{Image quality metrics for composition of objects onto held-out views in the real cars dataset. }
    \vspace{-0.0em}
    \label{tab:real-known}
    \vspace{-1em}
\end{table} 

%% file: tables/tab-comparison.tex
\begin{table}[h!]
\vspace{-0.5em}
\footnotesize
    \centering
    \begin{tabular}{l|cccc}
    
     & FID  $\downarrow$  & SSIM $\uparrow$ & PSNR $\uparrow$ & LPIPS $\downarrow$ \\
     \specialrule{0.12em}{0.05em}{0.05em}
    NeRF only & $0.452$ & $0.805$ & \tn{22.506}{2.26} & $0.177$  \\
    LANe (multiple) & $0.323$ & $0.866$ & \tn{26.49}{1.93} & $ 0.134 $  \\
    LANe (LF) & $0.145 $  & $ 0.875$ & \tn{25.056}{1.69} & $ 0.103 $  \\
    \hline
    \end{tabular}
    \vspace{-1mm}
    \caption{A comparison of image quality when composing the LANe-LF model into an unseen environment. Note that the other models have been trained on this environment.}
    \vspace{-0.0em}
    \label{tab:ablation_sg}
    \vspace{-1em}
\end{table}

%% file: sections/discussions.tex
\section{Discussions}
The dataset generated by CARLA exhibit spatially varying lighting effects due to both direct (sun) and indirect lighting (shadows cast by buildings). We see from Table.~\ref{tab:comparison} that our model is able to better capture changes in object lighting as it moves through the scene. Particularly, we see that a naive conditional baseline is insufficient as it correlates scene effects and object effects. Further more, naive conditioning requires large amount of data to accurate model the light transport across the scene.

Kundu et al. \cite{kundu2022panoptic} model each object with a seperate light MLP. using \texttt{FedAvg} to build in class priors. However, these model grows linearly in the number of objects being represented. We employ a variant of CodeNerf \cite{jang2021codenerf} to allow for larger degree of control over the class of objects without explosion in memory bandwidth.

%% file: sections/conclusion.tex
\section{Limitations and Conclusion}
\textbf{Limitation and future work} Our framework assumes observations of the object under varying lighting which limits its applicability to the scenes  where an object is seen under varying lighting conditions. This can be addressed using shading methods trained on both synthetic and real data. In addition,our approach does not model object-to-scene and object-to-object shadows. The shadows in real results are residuals from background. Addressing this would involves re-evaluating the lighting on an object once another is added, and is an exciting problem for future work, potentially using shadow fields for each instance that are dependent on the same lighting representation. Our model, much like many of the cited works, is sensitive to the pose parameters and object masks and pose robustness is outside the scope of our approach. However, recent works which are robust to camera pose (such as BARF \cite{BARF2021} or SPARF\cite{SPARF2022}) can potentially be leveraged to address this problem.

\textbf{Conclusion} In this work, we introduce an approach to leverage spatial lighting-aware NeRFs to build composable 3D scene representations. We separate the scene into object and world NeRFs and introduce a  multiplicative shading model to condition the object's appearance on the scene lighting. This allows our object models to be composed into new world models in a lighting-aware manner without retraining any object parameters.

%% file: supplm.tex
\crefname{section}{Sec.}{Secs.}
\Crefname{section}{Section}{Sections}
\Crefname{table}{Table}{Tables}
\crefname{table}{Tab.}{Tabs.}

\appendix

\twocolumn[
\centering
\large
\textbf{LANe - Lighting Aware Neural Fields for Compositional Scene Synthesis - Supplementary}
\vspace{1.5em} 
]

\section{Introduction}
For demo videos of 3D neural rendering and details about the ablation studies, please visit our project website: \url{ https://lane-composition.github.io}. We also intend to release code and datasets there.

\input{supplm_sections/architecture.tex}

\input{supplm_sections/dataset.tex}

\input{supplm_sections/qual_results}

\input{supplm_sections/ablations.tex}

\input{supplm_sections/relight_baseline}

\input{supplm_sections/limitations.tex}

\input{supplm_sections/societal_impact.tex}

%% file: supplm_sections/architecture.tex
\section{Architecture details}

The architectures for our world NeRF, world light field network, and object NeRFs are standard MLPs, optionally conditioned on latent codes to share models across instances, and are described in Sections \ref{sec:wm}, \ref{sec:om} and \ref{sec:arch}. Here we provide more details on our novel LFN-based shader network presented in Section \ref{sec:unknown}.

\subsection{Unknown scene - LFN based shader field}

Lighting-aware composition of neural fields in unknown scenes uses 3 components:

\begin{enumerate}[noitemsep]
    \item Object model: A lighting-agnostic object (car) model that was trained on images from different scenes. 
    \item Light field network for the scene to be composed into. 
    \item Shader network for the object to the composed.
\end{enumerate}

\input{figures/supplm/fig-light_field_arch.tex}

These components are illustrated in (Fig.~\ref{fig:supplm_light_field_arch}). 
In particular, the object model 
\begin{equation}
\mathbf{c},\sigma = \mathcal{N}_{obj}(\phi(\mathbf{x_{obj}}),\eta) 
\end{equation}
where, $\phi$ are the positional encoding and $\eta$ is the object code for the particular instance of the car model. For composing the objecting into scenes unseen during training, we use the light field of the new scene:
\begin{equation}
\mathbf{l_d} = \mathcal{N}_{light}(\mathbf{x},\mathbf{d})
\end{equation}
where $\mathbf{x} \in \mathbb{R}^3$ is the 3D location of a point on the object and $\mathbf{d} \in \mathbb{R}^3$ is the  direction of secondary ray, both expressed in world coordinates, and $\mathbf{l_d}$ is the incoming 3-channel LDR radiance at $\mathbf{x}$ along $\mathbf{d}$. 

To obtain the shading at a surface point $\mathbf{p}_{obj}$ in object coordinate frame, we first compute a local accumulated lighting feature using an attention mechanism:
\begin{equation}
    \mathbf{\tilde{f_l}} = \text{Attention}(QW_q, KW_k, VW_v)
\label{eq:attention}
\end{equation}

where $Q, K, V$ are queries, keys and values, and $W_q, W_k, W_v$ are their learnable linear mappings respectively. We use $Q= \phi(\mathbf{p}_{obj})$, and $K=V= \phi(\mathbf{d}_{obj}) \bigoplus \mathbf{l_d}$, where $\bigoplus$ is the concatenation operator and $\phi$ is the positional encoding. The attention mechanism has been leveraged to encourage the network to focus on some of all incoming light, conditioning on its local coordinates. This resembles the weighted integral of irradiance based on the incident angle in the rendering equation. Ideally, we expect the attention weights to be higher along the direction of the surface normal at the point. We visualize the attended direction at each surface point by obtaining the weighted average secondary direction using the learned attention weights (the ``attended direction'') in Fig. \ref{fig:attn_viz}. This shows that the attended direction aligns with the surface normals on the surface of a single car instance.

We then use a 4-layer MLP to predict a 3-channel shading coefficient for each point on the object conditioned on the learned accumulated lighting feature: 
\begin{equation}
    s_{car} = \mathcal{N}_{shading}(\mathbf{\tilde{f_l}},\mathbf{p}, \eta)
\end{equation}
where $\eta$ is the latent code for the specific instance. Note that we only train $\mathcal{N}_{shading}$ on points on the surface the object by thresholding points with an occupancy value greater than 0.5, to avoid sampling secondary rays at all points.

\input{figures/supplm/attn_viz.tex}

%% file: figures/supplm/fig-light_field_arch.tex
\begin{figure}[h!]
    \centering
    \includegraphics[width=\linewidth]{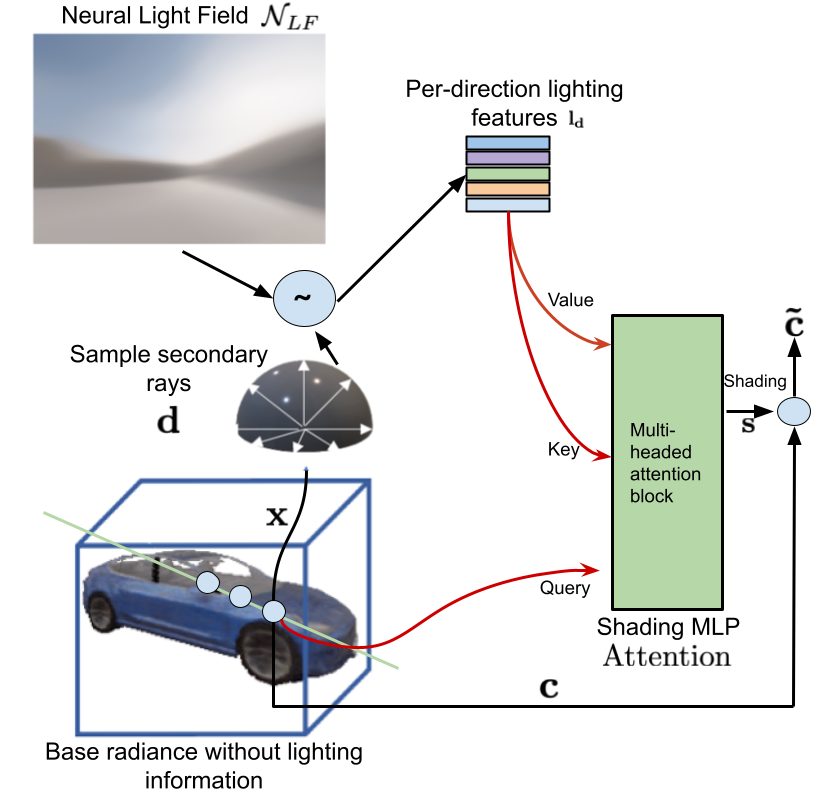}
    \caption{The architecture of the light field (LF) based shading network for compositing objects into unseen scenarios, which the models have never been trained with.}
    \label{fig:supplm_light_field_arch}
    \vspace{-0.3cm}
\end{figure}

%% file: figures/supplm/attn_viz.tex
\begin{figure}[h!]
    \centering
    \includegraphics[width=\linewidth]{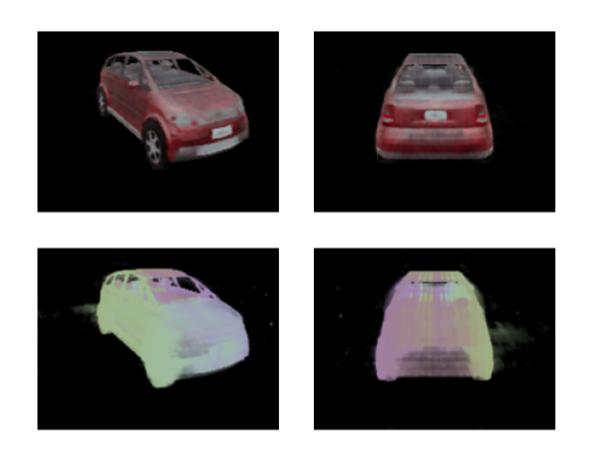}
    \caption{Direction obtained by the weighted average of all key attention weights. First row shows rendered RGB image, second row shows the 3D attended direction as RGB values. The attended directions tend to align with the surface normals.}
    \label{fig:attn_viz}
    \vspace{-0.3cm}
\end{figure}

%% file: supplm_sections/dataset.tex
\section{Dataset details}

In this section we describe and provide examples from the synthetic and real datasets used to train our models. 

\subsection{Synthetic dataset}

\input{figures/supplm/synthetic_dataset}

Our synthetic dataset comprises 6 different car instances rendered in 10 different lighting conditions using the CARLA simulator. Each lighting condition can be in one of 5 different scenes. Within each lighting condition, the car is observed in multiple positions within the scene, at different orientations. The camera viewpoint does not change significantly, but we still obtain images of different portions of the car as the car still has different poses with respect to the scene. This resembles the conditions encountered in real world self-driving datasets. We also use 3D bounding boxes and instance masks from the simulator. Examples from our dataset is illustrated in Fig \ref{fig:dataset_synth}.

\subsection{Real dataset}

\input{figures/supplm/real_dataset}
\input{figures/supplm/real_dataset_lf}

Our real dataset features sequences of images collected by moving a handheld camera around a parked vehicle. We collect sequences for the same vehicle at 10 different times of the day, for 3 different vehicles. Each vehicle is parked in a slightly different scene. At each time of day, we also collect an additional sequence with the camera zoomed out, so as to capture sky lighting and train a light field model with it. We use ~100 images from each sequence for training the object and shader models, and ~50 images for training the world light field models. We register all images for a particular instance into a common frame of reference using COLMAP \cite{schoenberger2016sfm}. The local frame for the vehicle is defined by manually annotating a 3D bounding box around the instance from the sparse COLMAP reconstruction. Instance masks are obtained from a pretrained Mask R-CNN \cite{maskrcnn2017he} model. Examples from the object model sequences are shown in Fig \ref{fig:dataset_real} and the light-field model sequences are shown in \ref{fig:dataset_real_lf}.

%% file: figures/supplm/synthetic_dataset.tex
\begin{figure}[h!]
    \centering
    \includegraphics[width=\linewidth]{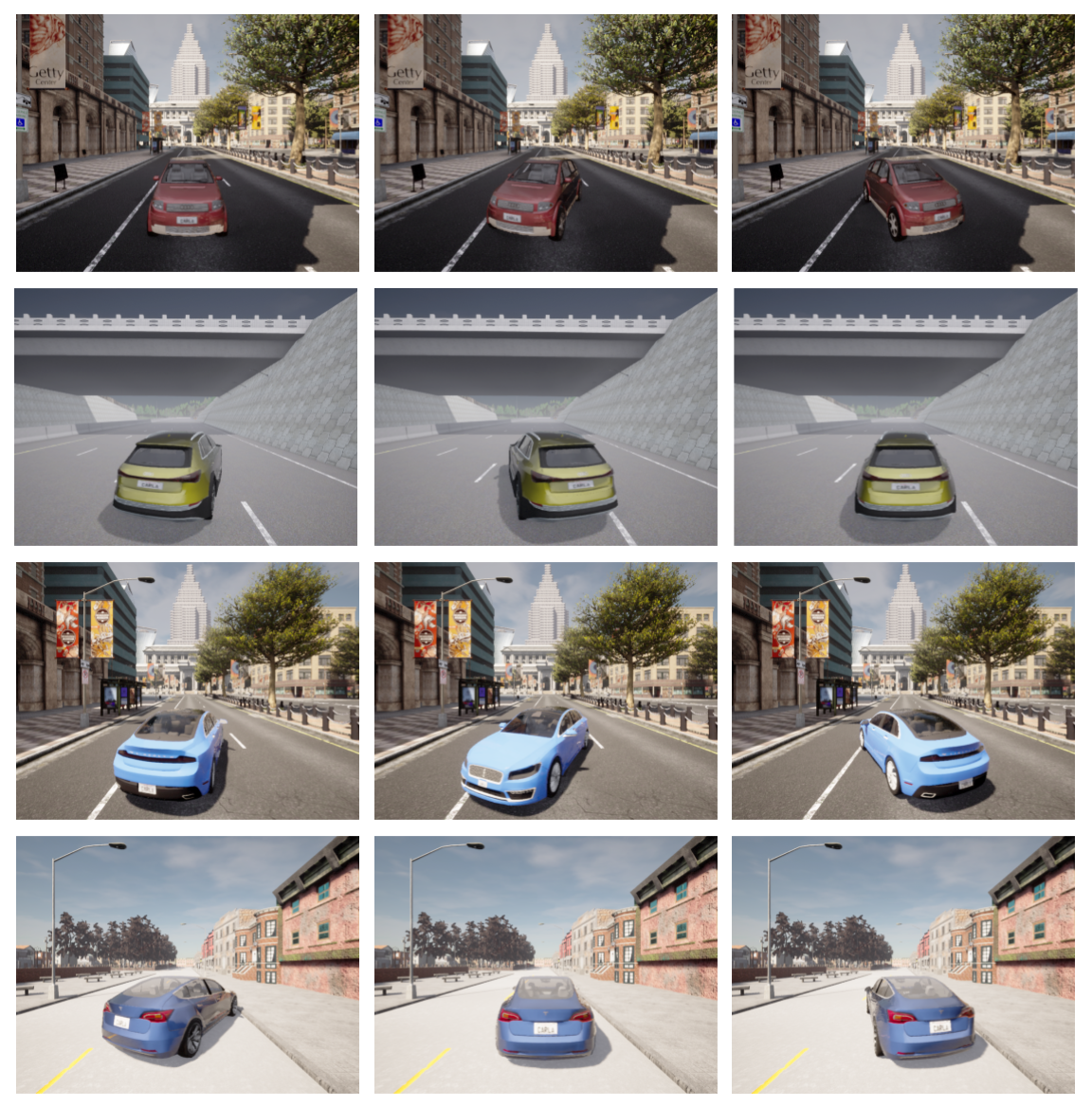}
    \caption{Examples of images from the synthetic dataset. It contains 6 cars in 10 different lighting conditions, at 40 different locations in each condition.}
    \label{fig:dataset_synth}
\end{figure}

%% file: figures/supplm/real_dataset.tex
\begin{figure}[h!]
    \centering
    \includegraphics[width=\linewidth]{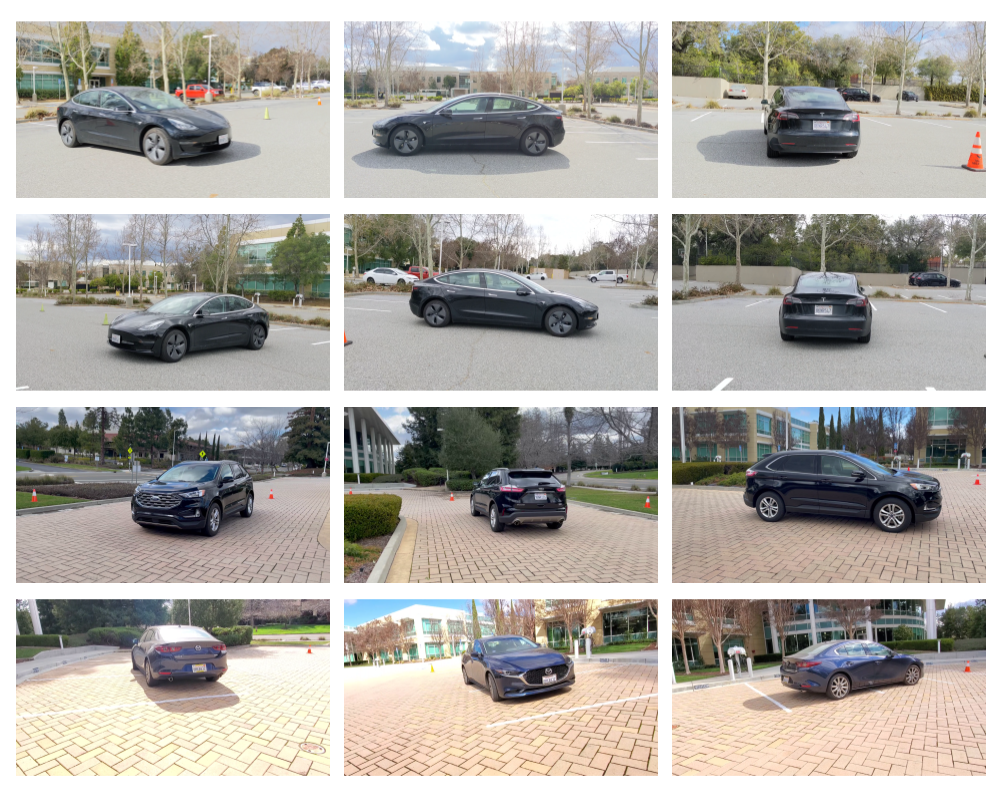}
    \caption{Examples of images from the real dataset collected. The dataset contains sequences of 3 cars at 10 different times of day.}
    \label{fig:dataset_real}
\end{figure}

%% file: figures/supplm/real_dataset_lf.tex
\begin{figure}[h!]
    \centering
    \includegraphics[width=\linewidth]{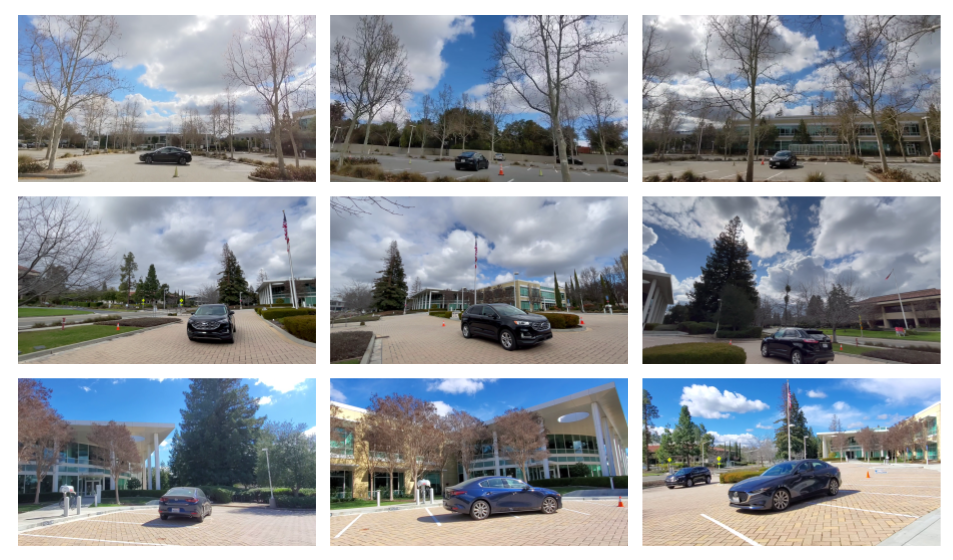}
    \caption{Examples of images from the real dataset collected for training the world light field models. We collect one such sequence per instance per time of day. Training the light field model requires large portions of the sky to be observed in these images.}
    \label{fig:dataset_real_lf}
\end{figure}

%% file: supplm_sections/qual_results.tex
\section{More qualitative results}

More qualitative results are provided for composing lighting-aware object models into both known and unknown scenes, for both synthetic (Fig \ref{fig:qual_synth}, \ref{fig:supplm_unknown_scenes}) and real datasets (Fig \ref{fig:qual_real}). Our approach also allows us to compose multiple objects into the same scene, as shown in Fig \ref{fig:multi_obj_synth}. 

For the results on real datasets in Fig \ref{fig:qual_real}, the object shadows in the composed image are part of the background image, not the object model. Further, as mentioned in \ref{sec:real-composition}, we do not use the attention layers for the shaders of the real-world object models, as these require more training data. We instead obtain the lighting feature $\mathbf{\tilde{f_l}}$ in Eq. \ref{eq:attention} from a global latent code used to condition the scene's light field network on the particular lighting ($\mathbf{f}$ in Eq. \ref{eq:world-light}). The latent-conditioned light field network is trained on the same training set as the shader, with a learned latent code per lighting condition. For unseen lighting conditions, the latent code is optimized by minimizing the light-field's reconstruction loss, and then used as an input for the shader. 

\input{figures/supplm/qual_synth}
\input{figures/supplm/qual_real}
\input{figures/supplm/mulit_obj_synth}

%% file: figures/supplm/qual_synth.tex
\begin{figure}[h!]
    \centering
    \includegraphics[width=\linewidth]{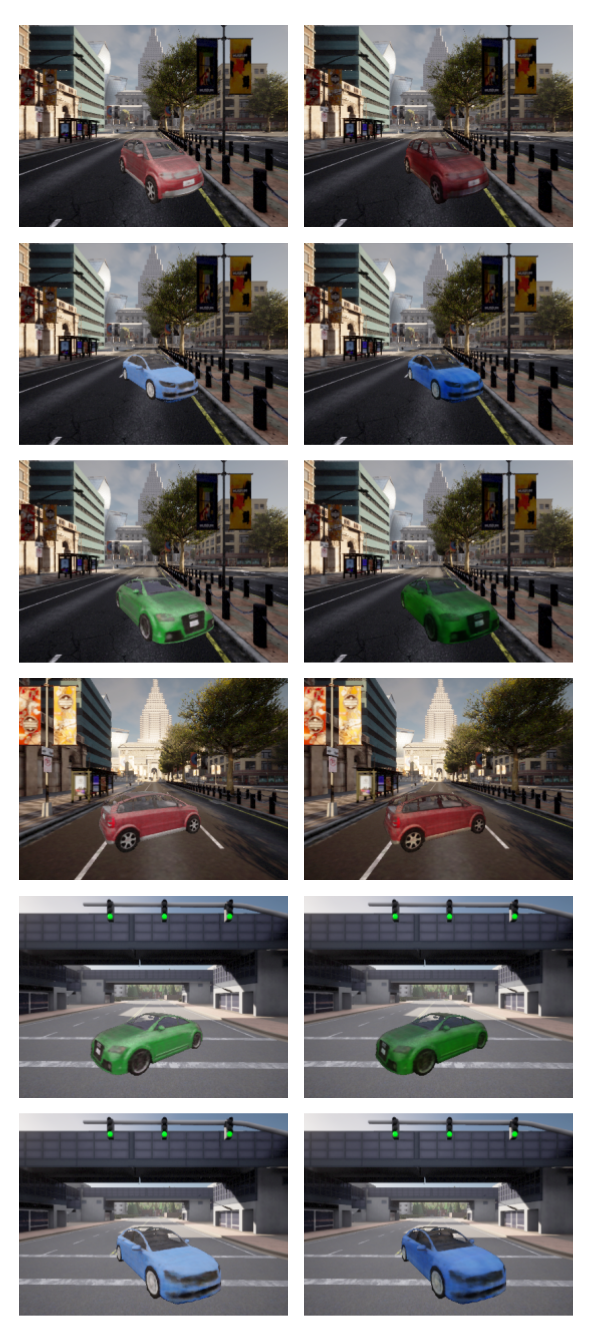}
    \caption{Examples of LANe shaded (right) and unshaded (left) object models composed onto different scenes. Note that the shader modulates the cars appearance differently to account for sun directions, shadows from adjacent buildings etc.}
    \label{fig:qual_synth}
\end{figure}

%% file: figures/supplm/qual_real.tex
\begin{figure}[h!]
    \centering
    \includegraphics[width=\linewidth]{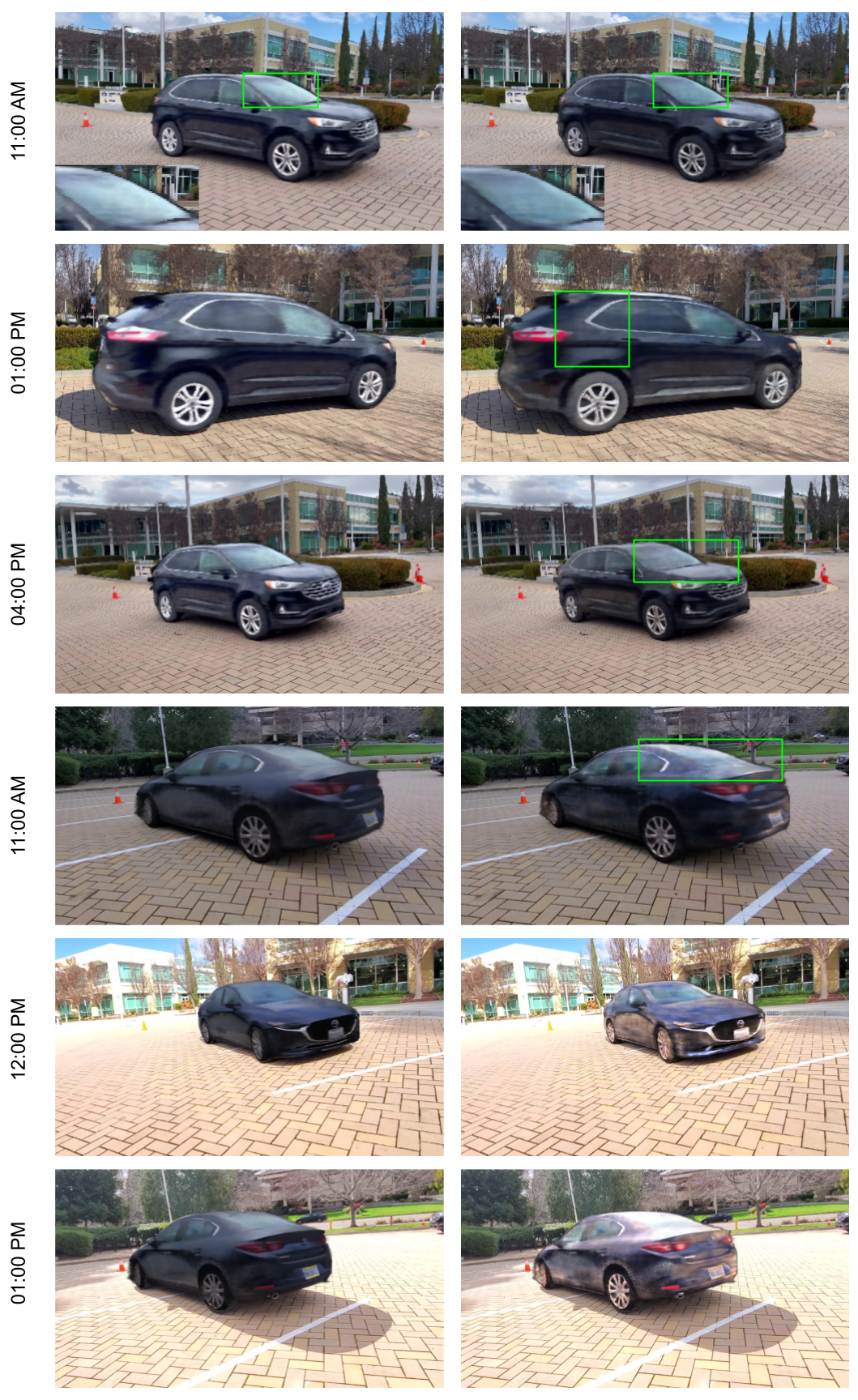}
    \caption{Examples of unshaded (left) and LANe shaded (right) object models composed onto the original positions of the car. The shader accounts for global changes in the scene lighting, darkening the cars in cloudy scenes and brightening them (with some specularities) in sunny scenes. In scenes where the shading changes are subtle, regions on the car with most change have been highlighted.}
    \label{fig:qual_real}
\vspace{-0.3cm}
\end{figure}

%% file: figures/supplm/mulit_obj_synth.tex
\begin{figure*}[h!]
    \centering
    \includegraphics[width=\textwidth]{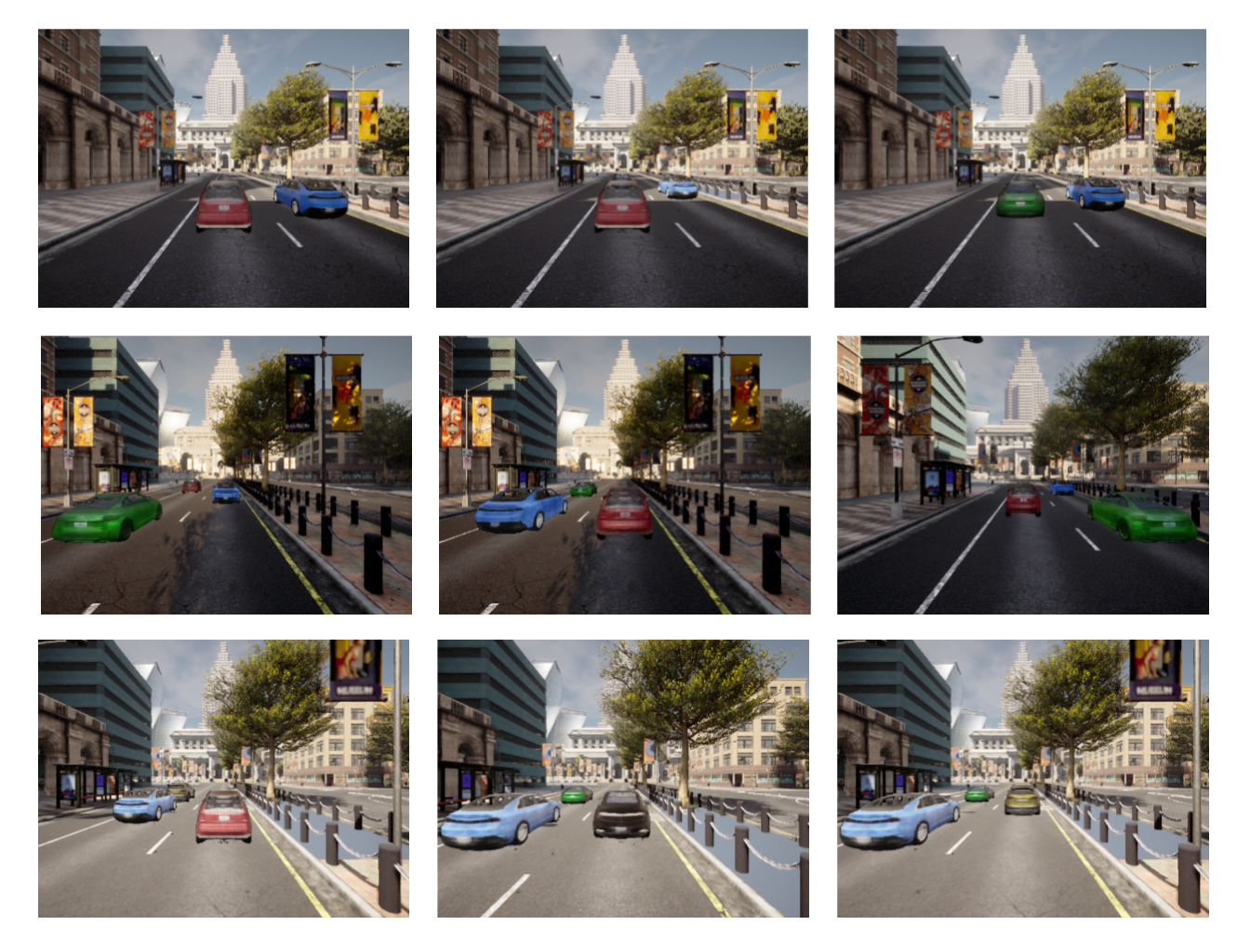}
    \caption{Composing multiple objects into seen and unseen lighting conditions using the attention based shader.}
    \label{fig:multi_obj_synth}
\end{figure*}

%% file: supplm_sections/ablations.tex
\section{Ablations}

The performance of the shader network depends on many architectural design choices. 
\input{figures/supplm/unknown_scenes.tex}

\subsection{Number of secondary rays}

The LANe architecture that generalizes to unseen scenes requires sampling secondary rays from points on the car to query the light field network for the environment. 

\input{figures/supplm/num_secondary.tex}

As indicated in Table.~\ref{tab:second_rays_seen_env},  the quality of the lighting-aware composition depends on the number of secondary rays sampled. It is natural to expect that the quality improves with an increase in the number of sampled secondary rays. We find that while this is indeed the case in environments observed during training, sampling fewer rays improves lighting-aware composition in unseen environments, as shown in Table ~\ref{tab:second_rays_unseen_env}. Our hypothesis is that the limited capacity of the attention layers causes them to learn an averaged representation of the light field when the number of rays increases. This representation is more biased to the overall lighting of the seen environments and is less sensitive to spatially varying lighting in the unseen environment. 

\begin{table}
\centering
\begin{tabular}{ccccc}
\specialrule{0.12em}{0.05em}{0.05em}
 Num. rays  &  PSNR $\uparrow$   &  SSIM $\uparrow$  &  FID $\downarrow$  &  LPIPS $\downarrow$ \\
\specialrule{0.12em}{0.05em}{0.05em}
       144        & \textcolor{blue}{\textbf{23.4441}} & \textcolor{blue}{\textbf{0.8495}} & \textcolor{blue}{\textbf{0.1742}} & 0.1027 \\
        72        & 23.0857 & 0.8467 & 0.1771 & 0.1018 \\
        54        & 22.9917 & 0.8441 & 0.1837 & 0.1038 \\
        36        & 22.8978 & 0.8448 & 0.2081 & 0.1006 \\
        24        & 22.7281 & 0.8437 & 0.2335 & 0.1005 \\
        18        & 22.6449 & 0.8430 & 0.2069 & \textcolor{blue}{\textbf{0.1004}} \\
\hline
\end{tabular}
\caption{Quality as a function of secondary rays for seen environments. The number of rays used for LANe has small impact on pixel-based image quality metrics (PSNR, SSIM) and learned perceptual similarity metric (LPIPS), but affects the FID score more (measuring difference between the ground truth and generated image distributions), shown more score variance with the decrease of number of ray used. The best trained model for seen environment has been obtained with the maximum of 144 rays setting}
\label{tab:second_rays_seen_env}
\end{table}

\begin{table}
\centering
\begin{tabular}{ccccc}
\specialrule{0.12em}{0.05em}{0.05em}
  Num. rays  &   PSNR $\uparrow$   &  SSIM $\uparrow$  &  FID $\downarrow$  &  LPIPS $\downarrow$ \\
\specialrule{0.12em}{0.05em}{0.05em}
       144        & 23.5663 & 0.8301 & 0.1321 & 0.1096 \\
        72        & 23.9249 & 0.8340 & 0.1327 & 0.1103 \\
        54        & 23.3989 & 0.8225 & 0.1362 & 0.1154 \\
        36        & 24.8212 & 0.8434 & 0.1136 & 0.1080 \\
        24        & \textcolor{blue}{\textbf{25.1234}} & \textcolor{blue}{\textbf{0.8439}} & \textcolor{blue}{\textbf{0.1072}} & 0.1089 \\
        18        & 24.6220 & 0.8435 & 0.1391 & \textcolor{blue}{\textbf{0.1017}} \\
\hline
\end{tabular}
\caption{Quality as a function of secondary rays for unseen environments. LANe model trained with fewer number of rays obtained better image composition quality. Overall, the model training setting of 24 rays achieved the best results for unseen environments.}
\label{tab:second_rays_unseen_env}
\end{table}

\subsection{Number of training environments and generalization}
\input{figures/supplm/num_instances.tex}
We seek to answer the question - how does increasing the number of training environments affect the performance of the LFN-based shader field on seen and unseen environments?  To this end, we train the shader fields on a varying number of environments - 3, 5, 7 - and evaluate them both on an environment used in all the training sets, and a completely unseen environment. We find that increasing the number of environments leads to a slight drop in performance on the training environment (Table ~\ref{tab:train-envs}), but a great increase in performance on the test environment (Table ~\ref{tab:test-envs}).  

\begin{table}[]
    \centering
        \begin{tabular}{ccccc}
        \hline
          Num. train envs  &  PSNR $\uparrow$   &  SSIM $\uparrow$  &  FID $\downarrow$  &  LPIPS $\downarrow$ \\
        \hline
                 3         & \textcolor{blue}{\textbf{22.2764}} & \textcolor{blue}{\textbf{0.8706}} & 0.2574 & 0.0975 \\
                 5         & 22.2113 & 0.8654 & \textcolor{blue}{\textbf{0.2274}} & 0.0990 \\
                 7         & 21.3136 & 0.8593 & 0.3635 & \textcolor{blue}{\textbf{0.0966}} \\
        \hline
        \end{tabular}
    \caption{Lighting aware reconstruction quality on a training scene when trained with an increasing number of environments. This shows that training in fewer environments can improve LFN-based shader's performance in those environments.}
    \label{tab:train-envs}
\end{table}

\begin{table}[]
    \centering
        \begin{tabular}{ccccc}
        \hline
          Num. train envs  &  PSNR $\uparrow$   &  SSIM $\uparrow$  &  FID $\downarrow$  &  LPIPS $\downarrow$ \\
        \hline
                 3         & 22.4900 & 0.8392 & 0.1551 & 0.1219 \\
                 5         & 22.8970 & 0.8514 & \textcolor{blue}{\textbf{0.1496}} & 0.1085 \\
                 7         & \textcolor{blue}{\textbf{24.0604}} & \textcolor{blue}{\textbf{0.8681}} & 0.1856 & \textcolor{blue}{\textbf{0.0968}} \\
        \hline
        \end{tabular}
    \caption{Lighting aware reconstruction quality on an unseen environment when trained with an increasing number of environments. This shows that training on more environments can improve LFN-based shader's generalizability. Note that the metrics on these unseen environments are comparable to those reported on a training scene.}
    \label{tab:test-envs}
\end{table}

\subsection{Number of instances}

The shared LANe object model uses a latent embedding to represent multiple car instances. We find that increasing the number of instances only slightly reduces the overall quality of its rendered images for the known-scene object model, keeping model capacity constant. The quality metrics as a function of the number of instances are shown in Table ~\ref{tab:of-instances}.

\begin{table}[]
    \centering
        \begin{tabular}{ccccc}
        \hline
          Num. instances  &  PSNR $\uparrow$   &  SSIM $\uparrow$  &  FID $\downarrow$  &  LPIPS $\downarrow$ \\
        \hline
                1         & \textcolor{blue}{\textbf{26.6484}} & \textcolor{blue}{\textbf{0.9501}} & \textcolor{blue}{\textbf{0.1229}} & \textcolor{blue}{\textbf{0.0553}} \\
                2         & 24.6540 & 0.8813 & 0.3993 & 0.1085 \\
                3         & 24.0274 & 0.8586 & 0.4580 & 0.1223 \\
                4         & 23.8273 & 0.8474 & 0.5053 & 0.1386 \\
                5         & 23.0509 & 0.8314 & 0.7595 & 0.1630 \\
                6         & 22.7355 & 0.8232 & 0.6089 & 0.1708 \\
        \hline
        \end{tabular}
    \caption{Lighting aware reconstruction quality with increasing number of instances, for a known-scene model. This does not change the number of training parameters in the model.}
    \label{tab:of-instances}
\end{table}

\subsection{Resolution of light field network}

The light field network has been trained to accommodate various lighting requirements using images of the scene at different positions and orientations (Fig \ref{fig:dataset_real_lf} for real datasets). The input to this network is a normalized position and ray direction, expressed in Plucker coordinates. The resolution of the light field network depends on the dimensionality of the cosine positional encoding used at the input to the network, higher frequency encodings provide more accurate light fields. When a positional encoding is not used, the light field is very smooth and blurry, and resembles a lighting map of the scene, as shown in Fig ~\ref{fig:supplm_lf}. We find that the PSNR of the LANe model obtained using light fields with and without position encodings are similar, but using  position encodings can sometimes result in periodic shading artifacts. We choose the low-frequency light fields without position encodings for our best-performing model, especially since the light field is queried only in sparse directions and we do not need the precise structure of the world in the lighting information.

\subsection{Regressing a multiplicative shading factor vs directly regressing RGB values}

The LANe model predicts a shading factor to condition appearance on lighting. An alternative is to directly predict an RGB value from the shader. We find that this performs equally well for single-instance LANe models (Table ~\ref{tab:direct_rgb}), and is therefore a viable alternative.

\begin{table}[]
    \centering
        \begin{tabular}{ccccc}
        \hline
          Architecture  &   PSNR $\uparrow$   &  SSIM $\uparrow$  &  FID $\downarrow$  &  LPIPS $\downarrow$ \\
        \hline
                Shading factor         & 23.4441 & 0.8495 & \textcolor{blue}{\textbf{0.1742}} & 0.1027 \\
                Color regression         & \textcolor{blue}{\textbf{23.4813}} & \textcolor{blue}{\textbf{0.8561}} & 0.2998 & \textcolor{blue}{\textbf{0.0957}} \\
        \hline
        \end{tabular}
    \caption{Reconstruction quality comparison between using a shading factor and directly regressing the color values.}
    \label{tab:direct_rgb}
\end{table}

\input{figures/supplm/shadow_residual.tex}
\input{figures/supplm/light_fields.tex}

\subsection{View-dependent shader}

Our shader MLP makes a Lambertian assumption and does not model view-dependent radiance. We evaluate whether including the viewing direction as another input to the shader improves shaded image quality for the real dataset. A comparison of view-dependent shading to view-independent shading is shown in Table \ref{tab:viewdir-shader}. We find that the shader is able to model view-dependent effects, and that the rendered image quality is slightly better than the one that does not use view-dependence. 

\begin{table}[]
    \centering
        \begin{tabular}{ccccc}
        \hline
          Model  &  PSNR $\uparrow$   &  SSIM $\uparrow$  &  LPIPS $\downarrow$ \\
        \hline
                View-dep. LANe (LF) & \textcolor{blue}{\textbf{27.7169}} & \textcolor{blue}{\textbf{0.9408}}  & 0.0582 \\
                LANE (LF)         & 26.6371 & 0.9374 & \textcolor{blue}{\textbf{0.0568}} \\
        \hline
        \end{tabular}
    \caption{Comparison of rendering quality for a view-dependent LANe (LF) shader to that of a viewing direction independent shader.}
    \label{tab:viewdir-shader}
\end{table}

%% file: figures/supplm/unknown_scenes.tex
\begin{figure*}[h!]
    \centering
    \includegraphics[width=\textwidth]{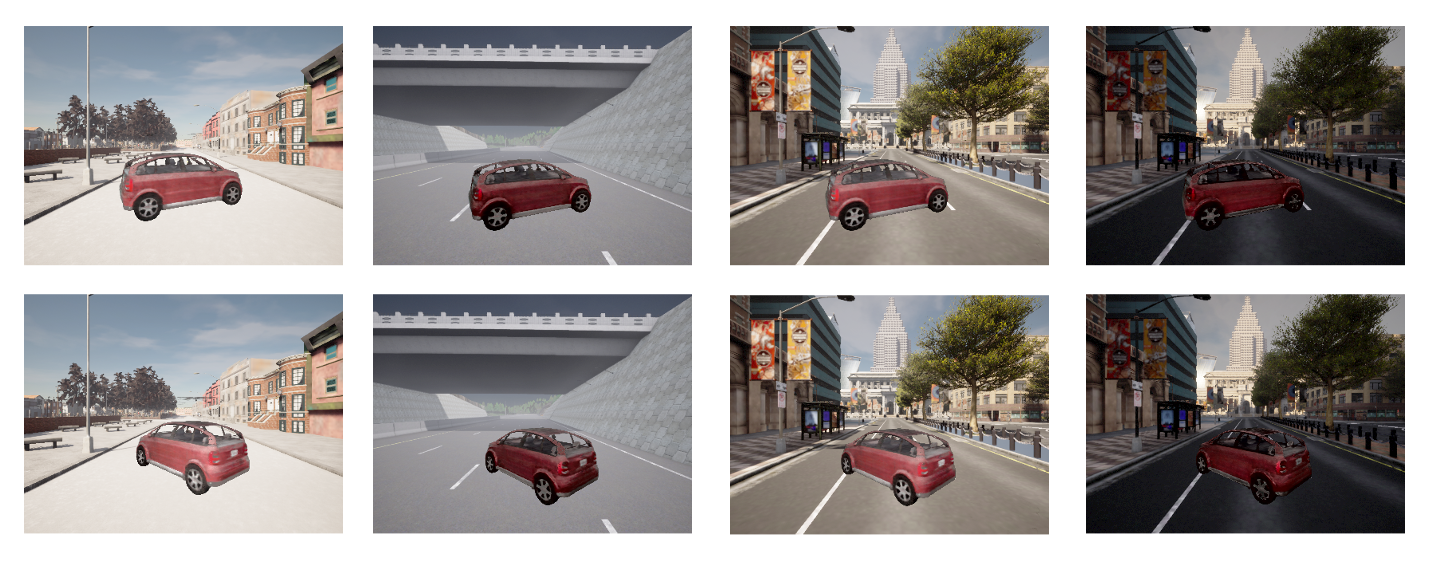}
    \caption{The LFN-based shader network can generalize to unseen environments and unseen lighting conditions in trained environments. This figure shows composing a car into some unseen environments.}
    \label{fig:supplm_unknown_scenes}
    \vspace{-0.3cm}
\end{figure*}

%% file: figures/supplm/num_secondary.tex
\begin{figure}[h!]
    \centering
    \includegraphics[width=\linewidth]{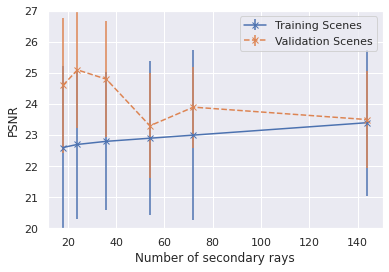}
    \caption{ Peak signal-to-noise ratio (PSNR) between training and validation scenes using different number of secondary rays.
    }
    \label{fig:supplm_num_secondary}
    \vspace{-0.4cm}
\end{figure}

%% file: figures/supplm/num_instances.tex
\begin{figure}[h!]
    \centering
    \includegraphics[width=\linewidth]{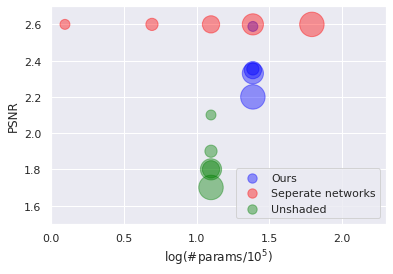}
    \caption{
    We show quality vs capacity tradeoff for out object model. The size of the points indicate the number of instances that can be represented by the model. We see that if we learn a seperate network for each object, the capacity quickly grows. However, for the same number of parameters our model can render a larger number of object instance for a slight drop in performance.
    }
    \label{fig:supplm_num_instances}
    \vspace{-0.3cm}
\end{figure}

%% file: figures/supplm/shadow_residual.tex
\begin{figure}[h!]
    \centering
    \includegraphics[width=\linewidth]{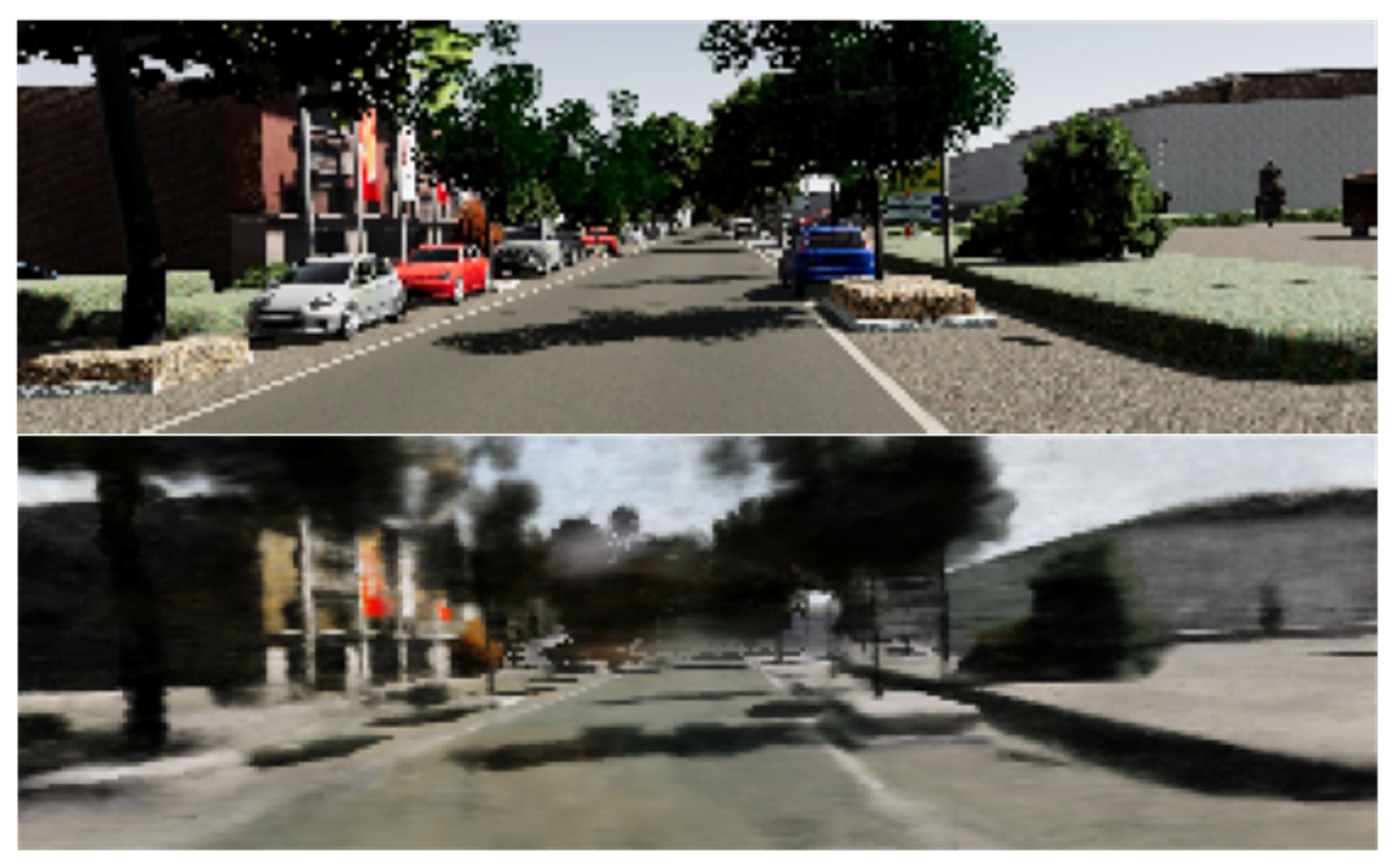}
    \caption{Top Row: Sample scene data consumed by LANe. Bottom Row: Background scene environment learnt and car object removed before new objects insertion, but hard cast shadow from the car objects remained}
    \label{fig:supplm_residual}
    \vspace{-0.3cm}
\end{figure}

%% file: figures/supplm/light_fields.tex
\begin{figure}[h!]
    \centering
    \includegraphics[width=\linewidth]{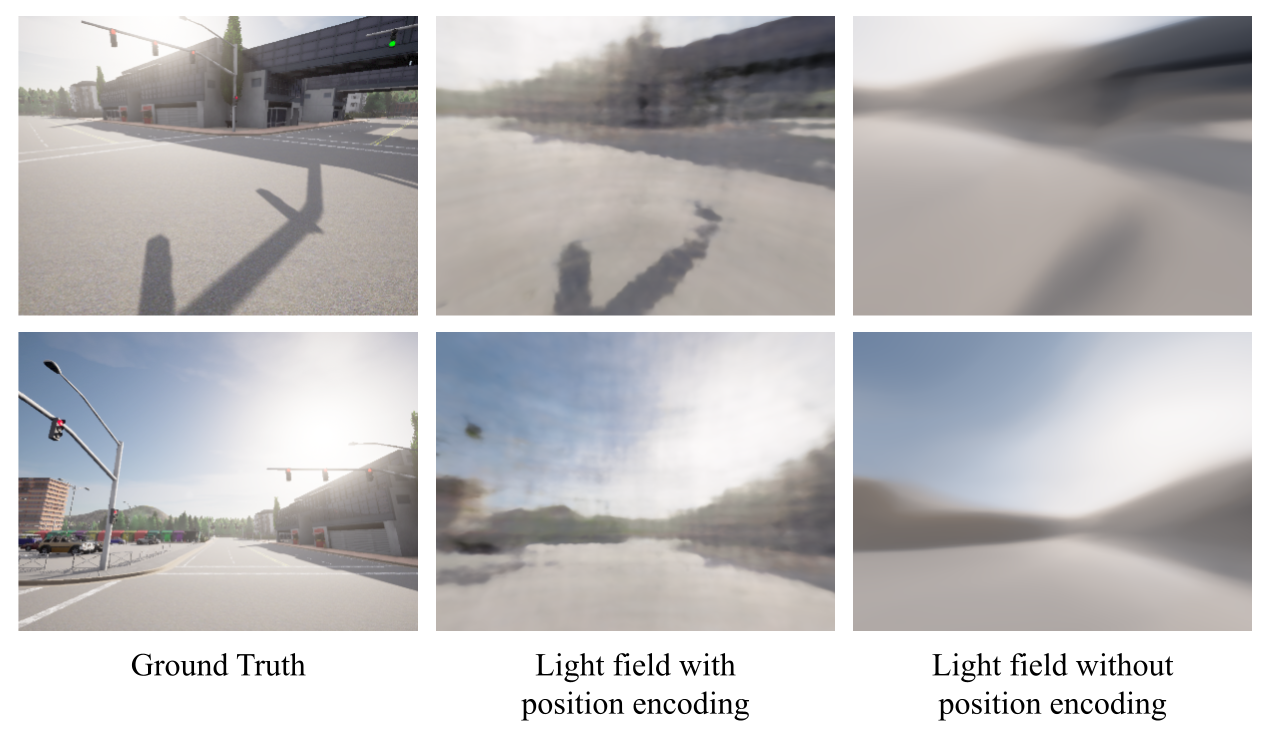}
    \caption{The resolution of light field and quality of LANe depends on the frequency of the input position embedding. We find that lower resolutions are more robust.}
    \label{fig:supplm_lf}
\end{figure}

%% file: supplm_sections/relight_baseline.tex
\section{Relightable NeRF baselines}

Several recent works \cite{boss2021nerd,srinivasan2021nerv,zhang2021nerfactor,boss2021neural, OSF2020} have addressed the problem of training relightable NeRF models by decomposing an object's radiance into its density, material (BRDF), albedo, lighting and visibility mask simultaneously. Such a decomposition is under-constrained and requires additional priors and regularizers. They also make assumptions on light-sources being at infinity, and do not model spatial variance within a scene. They have also not been demonstrated with noisy camera poses or on objects with transparent/translucent objects (such as window shields). Nevertheless, we attempted to compare against one lighting-aware baseline (NeRD \cite{boss2021nerd}). We found that this approach produces very  blurry results (Fig \ref{fig:supplm_nerd}) on our datasets where the object is not fixed relative to the environment.

\input{figures/supplm/nerd}

%% file: figures/supplm/nerd.tex
\begin{figure}[h!]
    \centering
    \includegraphics[width=\linewidth]{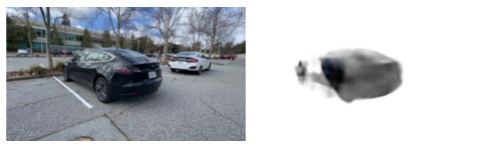}
    \caption{Our attempt to reproduce a relightable NeRF baseline (NeRD \cite{boss2021nerd}) produced blurry results (result on right, ground truth on left).}
    \label{fig:supplm_nerd}
    \vspace{-0.3cm}
\end{figure}

%% file: supplm_sections/limitations.tex
\section{Limitations}

\begin{enumerate}
\item Shadow modelling: The LANe model only modulates the appearance of the object with depending on the scene, and does not change the appearance of the scene or other objects after an object has been composed. This implies that effects on the scene (such as the composed object's shadow) is not modelled. 
\item Shadow residuals: As shown in Fig.~\ref{fig:supplm_residual}, it is challenging for LANe to cleanly remove existing hard cast shadow of the cars in the foreground, leaving some shadow residuals in the learned world model.
\item The lighting distribution between seen and unseen scenes have to be similar for composition into scenes that the object shader was not trained on. 
\item Data requirements: We assume that the same instance was visible in different lighting conditions to train the shader model. This is not true in data collected from real world driving scenarios, where each instance is only captured under lighting changes within the same scene. Training the multi-instance shader model jointly on synthetic instances rendered is several lighting conditions along with real instances observed in different positions in the same scene, could enable it to generalize to unseen real scenes. This is an interesting direction for future work. 
\end{enumerate}

%% file: supplm_sections/societal_impact.tex
\section{Societal Impact}

With the application of lighting-aware compositional scene synthesis using NeRF, LANe has great potential to be used for data augmentation to train  various downstream autonomous driving vision tasks. Specially, the learnt world model and object model could compromise an individual's privacy and safety, if it has been trained on images containing sensitive information. This is not a concern for the simulated data from CARLA used in our experiments. When releasing our real-world datasets or models trained on it, we intend to mitigate the privacy concern by not including any sensitive information, and  blurring information such as people and license plates in our images.